%% file: main.tex
\documentclass[10pt,twocolumn,letterpaper]{article}

\usepackage[pagenumbers]{cvpr} 


\definecolor{cvprblue}{rgb}{0.21,0.49,0.74}

\usepackage{array}
\usepackage{subcaption}
\usepackage{multirow,multicol}
\usepackage[flushleft]{threeparttable}
\usepackage{comment}
\usepackage{algorithmic}
\usepackage{booktabs}
\usepackage{bbm, dsfont}
\usepackage[font=small,labelfont=bf]{caption}

\usepackage{amssymb}
\usepackage{pifont}

\newcommand{\cmark}{\text{\ding{51}}}%
\newcommand{\xmark}{\text{\ding{55}}}%

\newcommand{\tablestyle}[2]{\setlength{\tabcolsep}{#1}\renewcommand{\arraystretch}{#2}\centering\footnotesize}

\newcolumntype{x}[1]{>{\centering\arraybackslash}p{#1pt}}
\newcommand{\app}{\raise.17ex\hbox{$\scriptstyle\sim$}}

\newlength\savewidth\newcommand\shline{\noalign{\global\savewidth\arrayrulewidth
  \global\arrayrulewidth 1pt}\hline\noalign{\global\arrayrulewidth\savewidth}}

\makeatletter\renewcommand\paragraph{\@startsection{paragraph}{4}{\z@}
  {.5em \@plus1ex \@minus.2ex}{-.5em}{\normalfont\normalsize\bfseries}}\makeatother

\def\tablecite#1#{%
  \def\pretablecite{#1}%
  \tableciteaux}
\def\tableciteaux#1{%
  \textsuperscript{\expandafter\originalcite\pretablecite{#1}}%
}

\usepackage{graphicx}
\usepackage{enumitem}
\usepackage{wrapfig}
\usepackage{lipsum}
\usepackage{soul}

\usepackage{color, colortbl}

\usepackage{tabu}
\usepackage{xcolor}
\usepackage{nicematrix}

\definecolor{ForestGreen}{rgb}{0.13, 0.55, 0.13}
\definecolor{Green}{rgb}{0.0, 0.5, 0.0}
\definecolor{Blue}{rgb}{0.25, 0.42, 0.88}
\definecolor{green(munsell)}{rgb}{0.0, 0.66, 0.47}
\definecolor{green(ryb)}{rgb}{0.4, 0.69, 0.2}
\definecolor{green(pigment)}{rgb}{0.0, 0.65, 0.31}
\definecolor{citecolor}{HTML}{0071bc}
\definecolor{GrayXMark}{gray}{0.7}
\definecolor{DifferenceColor}{HTML}{af3235}
\definecolor{HighlightColor}{gray}{0.9}
\definecolor{OracleTextColor}{gray}{0.55}
\definecolor{Cerulean}{HTML}{00a2e3}

\newcommand{\rownumber}[1]{\textcolor{Cerulean}{#1}}
\newcolumntype{H}{>{\setbox0=\hbox\bgroup}c<{\egroup}@{}}
\newcolumntype{a}{>{\columncolor{HighlightColor}}c}
\newcolumntype{L}[1]{>{\centering\arraybackslash}m{#1}}

\usepackage{tabularx}
\usepackage[export]{adjustbox}
\usepackage{makecell}
\usepackage{ragged2e}
\usepackage{amsmath}
\usepackage{amssymb}
\usepackage{pifont} 
\usepackage{enumitem}
\usepackage{booktabs}
\usepackage{setspace}
\usepackage[misc]{ifsym}

\newcommand{\ours}{ViLex\xspace}
\newcommand{\ourslc}{visual lexicon\xspace}
\newcommand{\ourfull}{Visual Lexicon\xspace}
\newcommand{\ourshort}{ViLex\xspace}
\newcommand{\dalle}{DALL\(\cdot\)E }

\newcommand{\hide}[1]{}

\definecolor{darkred}{rgb}{0.55, 0.0, 0.0}

\newcommand{\gxmark}{\textcolor{gray}{\xmark}}

\newcommand{\customfootnotetext}[2]{{
  \renewcommand{\thefootnote}{#1}
  \footnotetext[0]{#2}}}

\usepackage{dblfloatfix}

\usepackage[pagebackref,breaklinks,colorlinks,citecolor=cvprblue]{hyperref}

\usepackage[capitalize]{cleveref}
\crefname{section}{\S}{\S\S}
\crefname{subsection}{\S}{\S\S}
\crefformat{table}{Table~#2#1#3}
\crefformat{figure}{Figure~#2#1#3}
\crefformat{equation}{Eq~#2#1#3}

\def\paperID{4468} 
\def\confName{CVPR}
\def\confYear{2025}

\title{Visual Lexicon: Rich Image Features in Language Space}

\author{
  XuDong Wang$^{1,2*}$ \quad Xingyi Zhou$^{1}$ \quad Alireza Fathi$^{1}$ \quad Trevor Darrell$^{2}$ \quad Cordelia Schmid$^{1}$
   \\
   $^{1}$Google DeepMind  \quad \quad $^{2}$UC Berkeley 
}

\begin{document}

\twocolumn[{%
  \renewcommand\twocolumn[1][]{#1}%
  \maketitle
    \vspace{-12pt}
    \captionsetup{type=figure}
    \centering
    \includegraphics[width=1.0\textwidth]{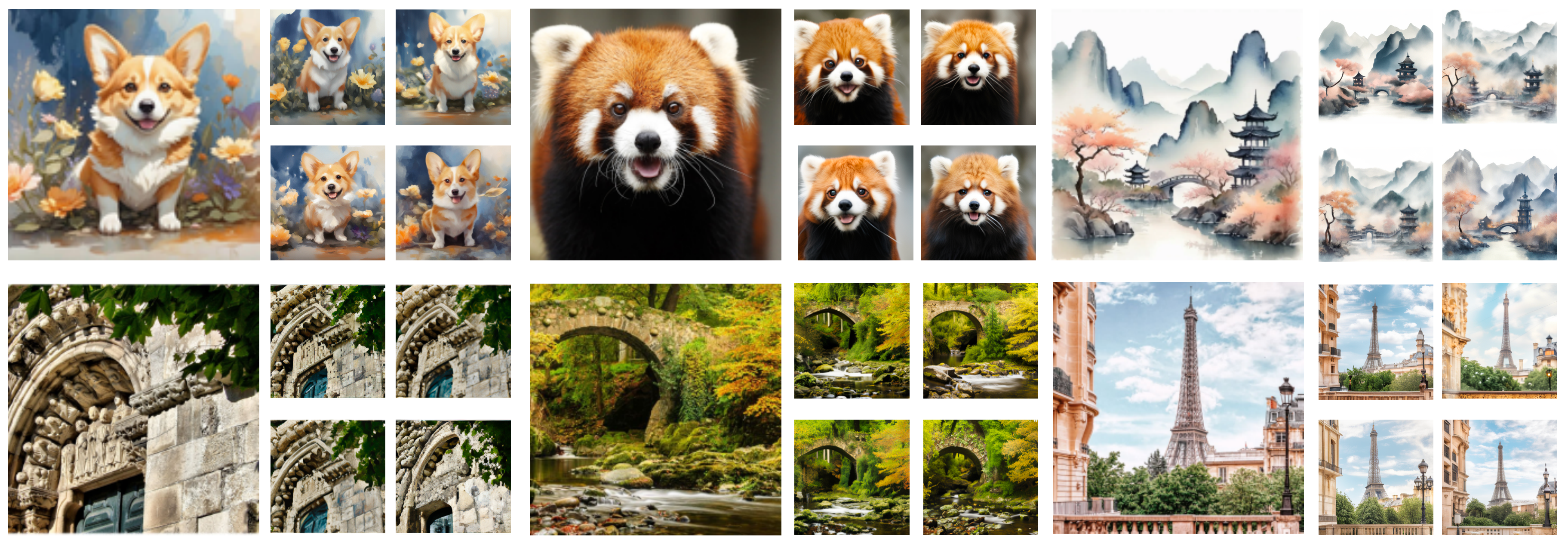}
    \vspace{-14pt}
    \caption{
 Given the cute corgi painting in the top left corner, how can we extract a visual representation that captures semantic-level information -- such as object categories and layouts -- while preserving rich visual details like image styles, textures and colors? \textbf{We introduce \ours model that generates image representations in the text vocabulary space}, 
 acting as a new visual ``language'', 
 while retaining intricate visual details that are difficult, if not impossible, to convey in natural language. 
 The set of images (generated under different diffusion noises) in the 2$\times$2 grid, which are highly semantically and visually similar to each other, is created by using \ours as ``text'' prompts for text-to-image diffusion models. 
    }
    \label{fig:teaser}
    \vspace{18pt}
}]
\customfootnotetext{}{\!\!\!\!\!\!\!\!\!\!\!\!*Work done in a Google DeepMind internship. \\ \Letter\{xudongwang, zhouxy\}\!@\!google.com\!\!}

\input{tables/vlm}
\input{tables/reconstruction}
\input{tables/ablation}

\input{sec/0_abstract}
\input{sec/1_intro}
\input{sec/2_related}
\input{sec/3_method}
\input{sec/4_experiment}
\input{sec/5_summary}


{
\small
\bibliographystyle{ieeenat_fullname}
\bibliography{main}
}

\input{sec/suppl}

\end{document}

%% file: tables/vlm.tex
\def\tabVQA#1{
    \begin{table}[#1]
    \tablestyle{0.5pt}{1.0}
    \small
    \begin{center}
    \begin{tabular}{lHcHlcHHccc}
    Model & Image Enc & \#toks & LLM &&
    \multicolumn{1}{c}{\rotatebox{90}{VQAv2}} & \multicolumn{1}{H}{\rotatebox{90}{TextVQA}} & \multicolumn{1}{H}{\rotatebox{90}{OKVQA}} & \multicolumn{1}{c}{\rotatebox{90}{SciQA}} & \multicolumn{1}{c}{\rotatebox{90}{VizWizQA}} & \multicolumn{1}{c}{\rotatebox{90}{GQA}} \\ [.1em]
    \Xhline{0.8pt}
    \hline
    \multicolumn{10}{c}{\bf \textit{Freeze image encoder}} \\
    PaliGemma-2B (SigLIP@224) & SigLIP@224 & 256 & 2B && 81.4 & 51.9 & 57.2 & 85.9 & 74.3 & 64.8 \\
    \rowcolor{gray!15}
    PaliGemma-2B (\ourshort\!@224) & \bf \ourshort\!@224 & 336 & 2B && \bf 83.6 & 59.4 & \bf 58.3 & \bf 88.6 & \bf 75.8 & \bf 65.7 \\
    \hline
    \multicolumn{10}{c}{\bf \textit{Fine-tune image encoder}} \\
    LLaVA1.5-7B (CLIP@224)\!~\cite{liu2024improved} & CLIP@224 & 576 & 7B && \phantom{1}78.5$^*$ && - & 66.8 & 50.0 & \phantom{1}62.0$^*$ \\
    LLaVA1.5-13B (CLIP@224)\!~\cite{liu2024improved} & CLIP@224 & 576 & 13B && \phantom{1}80.0$^*$ && - & 71.6 & 53.6 & \phantom{1}62.0$^*$  \\
    VILA-13B (CLIP@336)\!~\cite{lin2024vila} & CLIP@336 & 576 & 13B && \phantom{1}80.8$^*$ & \bf 64.4& - & 73.7 & 60.6 & \phantom{1}63.3$^*$ \\
    QwenVL-7B (CLIP@448)\!~\cite{bai2023qwen} & CLIP@448 & 1024 & 7B && \phantom{1}78.8$^*$ & \phantom{1}63.8$^*$ & 56.6 & 67.1 & 38.9 & \phantom{1}57.5$^*$ \\
    LLaVA1.5-13B (CLIP@336)$^\text{HD}$\!~\cite{liu2024improved} & CLIP@336 & 1280 & 13B && \phantom{1}81.8$^*$ & 62.5 & - & 71.0 & 57.5 & \phantom{1}63.3$^*$ \\
    LLaVA1.6-13B (CLIP@384)$^\text{HD}$\!~\cite{liu2024llava} & SigLIP@384 & 1280 & 7B && \phantom{1}81.8$^*$ && - & 70.2 &-& \phantom{1}64.2$^*$ \\
    \end{tabular}
    \end{center}\vspace{-20pt}
    \caption{
    \textbf{\ours can be a strong vision encoder for vision-language tasks}.
    Using the \textit{lowest image resolution}, \textit{fewer tokens} per image, \textit{the smallest LLM model}, and \textit{without fine-tuning image encoder}, our \ours\!—integrated into PaliGemma as the image encoder—achieves SOTA performance across multiple visual question answering tasks.
    *: The training images from the datasets are utilized during model training or for fine-tuning the model.}
    \label{tab:vqa}
    \end{table}
}


\def\tabMainCompare#1{
    \begin{table}[#1]
    \tablestyle{2pt}{1.0}
    \small
    \begin{center}
    \begin{tabular}{HlHHcHHcHHHcccHHHcHHccHH}
    Model & Image Encoder & \#Tokens & LLM & \multicolumn{1}{c}{\rotatebox{90}{COCOcap}} & \multicolumn{1}{H}{\rotatebox{90}{XM3600 (en)}} & \multicolumn{1}{H}{\rotatebox{90}{COCO-35L}} & \multicolumn{1}{c}{\rotatebox{90}{TextCaps}} & \multicolumn{1}{H}{\rotatebox{90}{SciCap}} &&
    \multicolumn{1}{H}{\rotatebox{90}{VQAv2}} & \multicolumn{1}{c}{\rotatebox{90}{TextVQA}} & \multicolumn{1}{c}{\rotatebox{90}{OKVQA}} & \multicolumn{1}{c}{\rotatebox{90}{SciQA}} & \multicolumn{1}{H}{\rotatebox{90}{VizWizQA}} & \multicolumn{1}{H}{\rotatebox{90}{GQA}} &&
    \rotatebox{90}{RC-val} & \rotatebox{90}{RC-testA} & \rotatebox{90}{RC-testB} & \rotatebox{90}{RCp-testA} & \rotatebox{90}{RCg-test} &&
    \multicolumn{1}{H}{\rotatebox{90}{MSRVTT}}
    \\ [.1em]
    \Xhline{0.8pt}
    \hline
    PaliGemma & VAE             & 256 & 2B & 61.4  &      &       & 15.0  & - & - & - & 12.9 & 33.3 & 82.1 & - & - & - & 19.5 & - & 20.6 & 12.1 & 13.8 & - & - \\
    \rowcolor{gray!15}
    \rowcolor{gray!15}
    PaliGemma & \ours SigLIP & 224 & 2B & \bf 141.5 & \bf 81.2 & \bf 139.6 & \bf 124.0 & \bf 159.7 && \bf 82.5 & \bf 52.9 & \bf 57.5 & \bf 87.9 & \bf 75.8 & \bf 65.7 && \bf 67.2 & \bf 73.0 & \bf 65.1 &\bf 65.3 & \bf 62.6 &&\bf 70.7  \\
    \rowcolor{gray!15}
    PaliGemma & \ours & 336 & 2B & \bf 142.8 & \bf 81.2 & \bf 139.6 & \bf 137.7 & \bf 159.7 && \bf 82.5 & \bf 59.4 & \bf 58.3 & \bf 88.6 & \bf 75.8 & \bf 65.7 && \bf 70.9 & \bf 73.0 & \bf 68.0 &\bf 69.2 & \bf 65.9 &&\bf 71.2  \\
    \end{tabular}
    \end{center}\vspace{-16pt}
    \caption{
    \textbf{Improving vision-language models with \ours tokens}. 
    Concatenating \ours tokens with SigLIP patch tokens enhances the performance of vision-language models across diverse downstream tasks, such as image captioning, visual question answering, and image segmentation, with a modest token increase of only 25\% (from 256 to 336).
    \textbf{Pixel-reconstruction or semantic-reconstruction?} Although VAE can preserve pixel-level details during image reconstruction, its lack of semantic richness results in significantly poorer performance in vision-language modeling compared to \ours.
    We use PaliGemma as the base VLM model. 
    }
    \label{tab:res-concat}
    \end{table}
}

\def\tabVEncoders#1{
    \begin{table}[#1]
    \tablestyle{2pt}{1.02}
    \small
    \begin{center}
    \begin{tabular}{llHHlccHcc}
    Model & Vision Encoder & Token & LLM &&
    \multicolumn{1}{c}{\rotatebox{90}{CoCoCap}} & \multicolumn{1}{c}{\rotatebox{90}{XM3600(en)}} & \multicolumn{1}{H}{\rotatebox{90}{CoCo-35L}} & \multicolumn{1}{c}{\rotatebox{90}{TextCaps}} & \multicolumn{1}{c}{\rotatebox{90}{SciCap}} \\ [.1em]
    \Xhline{0.8pt}
    \hline
    PaliGemma & SigLIP & 256 & 2B && 140.4 & 78.0 & 137.9 & 130.4 & 135.5 \\
    PaliGemma & VAE & 336 & 2B &&  61.4 & 57.2 & - & 15.0 & 90.7 \\
    \rowcolor{gray!15}
    PaliGemma & \bf \ours & 336 & 2B && \bf 142.8 & \bf 81.2 & \bf 139.6 & \bf 137.7 & \bf 159.7 \\
    \end{tabular}
    \end{center}\vspace{-18pt}
    \caption{
    Comparing the performance of various image encoders, including \ours, SigLIP, CoCa and VAE, on multiple image captioning tasks.}
    \label{tab:vqa}
    \end{table}
}

					


\def\tabSigLIP#1{
    \begin{table*}[#1]
    \tablestyle{2.1pt}{1.012}
    \small
    \begin{center}
    \begin{tabular}{lHccccclcccccclcccccclcc}
    &&& \multicolumn{5}{c}{\text{Image Captioning}} && \multicolumn{6}{c}{{Visual Question Answering}} && \multicolumn{6}{c}{Image Segmentation} && Video \\
    \cline{4-8} \cline{10-15} \cline{17-22} \cline{24-24}
    Backbone & Fine-tuned & FID ($\downarrow$) & {\rotatebox{80}{COCOcap}} & {\rotatebox{80}{COCO-35L}} & {\rotatebox{80}{TextCap}} & {\rotatebox{80}{SciCap-Val}} & {\rotatebox{80}{SciCap-Test}} &&
    {\rotatebox{80}{VQAv2-Val}} & \rotatebox{80}{TextVQA} & {\rotatebox{80}{OKVQA}} & {\rotatebox{80}{SciQA}} & {\rotatebox{80}{VizWizQA}} & {\rotatebox{80}{GQA}} &&
    \rotatebox{80}{RC-val} & \rotatebox{80}{RC-testA} & \rotatebox{80}{RC-testB} & \rotatebox{80}{RCp-testA} & \rotatebox{80}{RCp-testB} & \rotatebox{80}{RCg-test} &&
    {\rotatebox{80}{MSRVTT}} \\ [.1em]
    \Xhline{0.8pt}
    \hline
    Original SigLIP & \xmark &    2.54 &    139.7 &    138.6 &    122.1 &    131.7 &    135.5 && 81.4 &    51.9 &    57.1 &    85.9 &    74.3 &    64.8 &&    66.2 &    69.0 &    63.6 &    63.3 &    55.3 &    59.6 &&    69.4 \\
    \rowcolor{gray!15}
    \ours SigLIP    & \cmark &\bf 2.38 &\bf 141.5 &\bf 139.4 &\bf 124.0 &\bf 134.3 &\bf 136.2 &&\bf 81.6    &\bf 52.9 &\bf 58.4 &\bf 87.9 &\bf 74.9 &\bf 65.3 &&\bf 67.6 &\bf 70.0 &\bf 65.1 &\bf 65.3 &\bf 57.3 &\bf 62.6 && \bf 70.7 
    \end{tabular}
    \end{center}\vspace{-18pt}
    \caption{
    \textbf{\ours improves both image understanding and reconstruction capabilities of vision encoders} by fine-tuning them using \ours's training approach. 
    Compared with the official SigLIP model~\cite{zhai2023sigmoid}, \ours SigLIP, fine-tuned with \ours approach, demonstrates superior image reconstruction quality (evidenced by a lower FID score) and enhanced visual scene understanding (as shown by improved results on numerous vision-language tasks).  
    We utilize PaliGemma's~\cite{beyer2024paligemma} framework for linear evaluation, replacing the vision encoder with either the fine-tuned SigLIP in \ours or the official one, and freeze vision encoder and fine-tune the model on downstream tasks.  
    We use the same hyper-parameters and model architecture for a fair comparison.
    RC refers to RefCOCO dataset.}
    \label{tab:siglip}
    \end{table*}
}

%% file: tables/reconstruction.tex
\def\tabFID#1{
    \begin{table}[#1]
    \tablestyle{4.2pt}{1.0}
    \small
    \begin{center}
    \begin{tabular}{lcccccccc}
     & \multirow{2}{*}{T2I} & \multirow{2}{*}{DeDiffusion}  & \multicolumn{4}{c}{\ours} \\
    \cline{4-7} 
     & & & 1 & 4 & 16 & 75 \\ [.1em]
    \Xhline{0.8pt}
    FID ($\downarrow$) & 6.52 & 3.89    & 3.65  & 2.91  & 2.38  & \bf 2.07 \\
    IS  ($\uparrow$)   & 14.06 & 14.68  & 15.33 & 15.42 & 15.51 & \bf 15.88 \\
    \end{tabular}
    \end{center}\vspace{-18pt}
    \caption{Even when only using just one continuous token, \ours outperform the discrete tokens from both DeDiffusion~\cite{wei2024diffusion} (image$\rightarrow$text$\rightarrow$image) and the vanilla Imagen~\cite{saharia2022photorealistic} (text$\rightarrow$image). 
    FID scores on MSCOCO-64×64 were used for image reconstruction comparison across various image generation pipelines, all of which employ Imagen~\cite{saharia2022photorealistic} as the base text-to-image diffusion model for fair comparisons. IS refers to inception score.}
    \label{tab:fid}
    \end{table}
}

\def\tabHumanEval#1{
    \begin{table}[#1]
    \vspace{-4pt}
    \tablestyle{2.7pt}{1.0}
    \small
    \begin{center}
    \begin{tabular}{lccclccc}
    & \multicolumn{3}{c}{DeDiffusion} && \multicolumn{3}{c}{\dalle3} \\
    \cline{2-4} \cline{6-8}
    & layout & semantic & style && layout & semantic & style \\
    \Xhline{0.8pt}
    \textit{vs.} \ours ($\uparrow$) & 98\% & 95\% & 98\% && 91\% & 76\% & 90\% \\
    \end{tabular}
    \end{center}\vspace{-18pt}
    \caption{\textbf{Human studies} on generating semantically and visually similar images using the image-to-image pipeline. We report the percentage of ratings favoring \ours over DeDiffusion~\cite{wei2024diffusion} and the image-guided \dalle3~\cite{betker2023improving} in terms of layout, semantic, and style consistency with the input image.}
    \label{tab:humam-eval}
    \end{table}
}

%% file: tables/ablation.tex
\def\tabContribution#1{
    \begin{table}[#1]
    \begin{center}
    \vspace{-8pt}
    \tablestyle{4pt}{1.0}
    \small
    \begin{tabular}{cccccc|lll}
    \rownumber{\#} & {\rotatebox{80}{Comp1}} & {\rotatebox{80}{Comp2}} & {\rotatebox{80}{Comp3}} & {\rotatebox{80}{Comp4}} & {\rotatebox{80}{Comp5}} & {FID} & {COCOCaps} & {RefCOCO} \\
    \shline
    \rowcolor{HighlightColor}
    \rownumber{1} & \cmark & \cmark & \cmark & \cmark & \cmark &  \\
    \rownumber{2} & \gxmark & \cmark & \cmark & \cmark & \cmark & \\
    \rownumber{3} &\cmark & \gxmark & \cmark & \cmark & \cmark & \\
    \rownumber{4} & \cmark & \cmark & \gxmark & \cmark & \cmark & \\
    \rownumber{5} &\cmark & \cmark & \cmark & \gxmark & \cmark & \\
    \rownumber{6} &\cmark& \cmark & \cmark & \cmark & \gxmark & \\
    \end{tabular}
    \end{center}\vspace{-6pt}
    \caption{\textbf{Contribution of each component} evaluated by removing or adding it and measuring the impact of the generated image in terms of FID score and the performance of vision-language tasks such as image captioning and segmentation.
    }
    \label{tab:abl-components}
    \end{table}
}

\def\tabVariousEnc#1{
    \begin{table}[#1]
    \tablestyle{2pt}{1.02}
    \small
    \begin{center}
    \begin{tabular}{llHHlccHcc}
    Model & \multicolumn{2}{c}{CoCa} & \multicolumn{2}{c}{SigLIP} & \\
    
    \multicolumn{1}{c}{\rotatebox{90}{CoCoCap}} & \multicolumn{1}{c}{\rotatebox{90}{XM3600(en)}} & \multicolumn{1}{H}{\rotatebox{90}{CoCo-35L}} & \multicolumn{1}{c}{\rotatebox{90}{TextCaps}} & \multicolumn{1}{c}{\rotatebox{90}{SciCap}} \\ [.1em]
    \Xhline{0.8pt}
    \hline
    PaliGemma & SigLIP & 256 & 2B && 140.4 & 78.0 & 137.9 & 130.4 & 135.5 \\
    PaliGemma & VAE & 336 & 2B &&  61.4 & 57.2 & - & 15.0 & 90.7 \\
    \rowcolor{gray!15}
    PaliGemma & \bf \ours & 336 & 2B && \bf 142.8 & \bf 81.2 & \bf 139.6 & \bf 137.7 & \bf 159.7 \\
    \end{tabular}
    \end{center}\vspace{-18pt}
    \caption{
    \ours training pipeline can consistently improve the visual scene understanding capabilities of various vision encoders. We use PaliGemma framework and replace vision encoders with either a baseline encoder or the fine-tuned one with our \ours approach.}
    \label{tab:various-encoders}
    \end{table}
}

%% file: sec/0_abstract.tex
\begin{abstract}
We present \ourfull, a novel visual language that encodes rich image information into the text space of vocabulary tokens while retaining intricate visual details that are often challenging to convey in natural language.
Unlike traditional methods that prioritize either high-level semantics (e.g., CLIP) or pixel-level reconstruction (e.g., VAE), \ours simultaneously captures rich semantic content and fine visual details, enabling high-quality image generation and comprehensive visual scene understanding.
Through a self-supervised learning pipeline, \ours generates tokens optimized for reconstructing input images using a frozen text-to-image (T2I) diffusion model, preserving the detailed information necessary for high-fidelity semantic-level reconstruction.
As an image embedding in the language space, \ours tokens leverage the compositionality of natural languages, allowing them to be used independently as ``text tokens'' or combined with natural language tokens to prompt pretrained T2I models with both visual and textual inputs, mirroring how we interact with vision-language models (VLMs).
Experiments demonstrate that \ours achieves higher fidelity in image reconstruction compared to text embeddings—even with a single \ours token.
Moreover, \ours successfully performs various DreamBooth tasks in a zero-shot, unsupervised manner without fine-tuning T2I models.
Additionally, \ours serves as a powerful vision encoder, consistently improving vision-language model performance across 15 benchmarks relative to a strong SigLIP baseline.

\end{abstract}

%% file: sec/1_intro.tex
\def\figPipelineSimple#1{
    \captionsetup[sub]{font=small}
    \begin{figure}[#1]
    \setlength{\belowcaptionskip}{-4pt}
      \centering
      \includegraphics[width=1.0\linewidth]{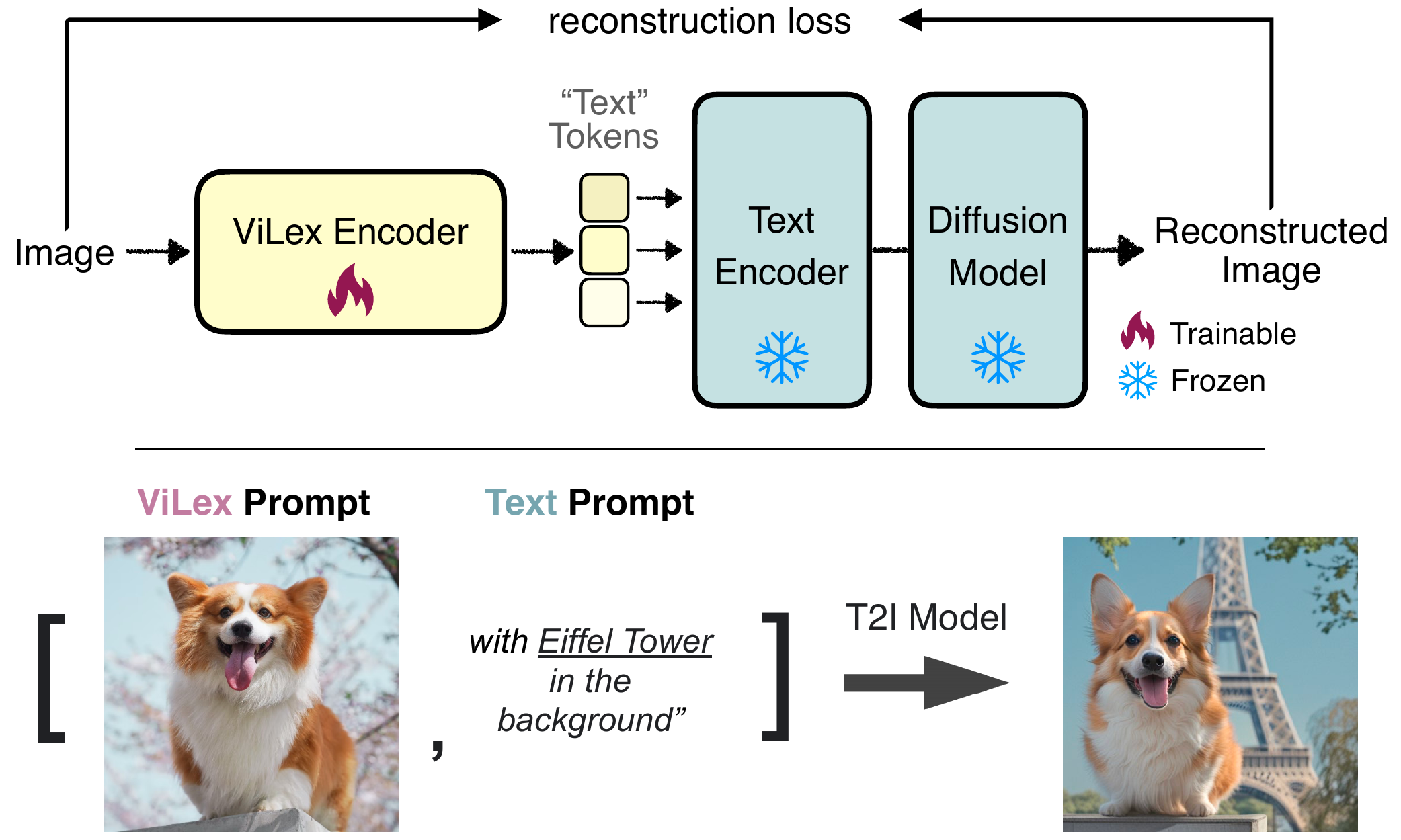}
      \vspace{-12pt}
      \caption{
      \textbf{top)} \textbf{ViLex empowers linguistic space to capture visual richness}. 
      We propose \textbf{\ours}, an image encoder that maps images into the vocabulary space, effectively preserving semantic information and intricate visual details.
      The embeddings from \ours function as a \textit{\ourfull} that preserve semantic  and intricate visual details of the image.
      \ours is trained with a frozen text-to-image diffusion model and can be utilized independently as ``text'' tokens for image generation. 
      \textbf{bottom)} \textbf{Linguistic space empowers ViLex to enjoy compositionality}.
      \ours can be combined with natural language tokens for prompting a pretrained T2I diffusion models with both visual and textual cues.
      }
      \vspace{-1mm}
      \label{fig:visualwords-simple}
    \end{figure}
}

\section{Introduction}
\label{sec:intro}

How should we represent an image? This is a fundamental question in computer vision. 
Over decades of progress, there have been two primary approaches: 
representations optimized for understanding high-level semantics~\cite{deng2009imagenet,radford2021learning,oquab2023dinov2}, or for high-fidelity image reconstruction~\cite{he2022masked,kingma2013auto,yu2024image}, often used in image generation~\cite{saharia2022photorealistic,ramesh2022hierarchical}. 
Understanding-focused models like CLIP~\cite{radford2021learning} and DINO~\cite{caron2021emerging} capture high-level semantics but lose pixel-level details. 
Conversely, reconstruction-focused models, such as VAEs~\cite{kingma2013auto}, retain fine visual details but lack semantic richness, making them less effective for tasks like vision-language modeling. 
In this paper, we aim to address the question: \textit{``can a single representation excel in both image reconstruction and semantic understanding?''}

\figPipelineSimple{t!}

To bridge this gap, we introduce \textbf{\ours} that encodes images into a {\textbf{Vi}sual \textbf{Lex}icon} within the text space. 
\ours model is designed to capture both high-level semantics -- such as object categories and spatial layouts -- while preserving rich visual details like styles, and textures that are difficult or even impossible to articulate in natural language.

We achieve this, as illustrated in \cref{fig:visualwords-simple}, by leveraging a self-supervised learning strategy based on a frozen, pretrained text-to-image (T2I) diffusion model, which acts as the source of supervisory signals.  
Although initially developed for generative purposes, many recent works~\cite{baranchuk2022labelefficient,li2023your,gan2024instructcv,xu2023open} have discovered that diffusion models~\cite{saharia2022photorealistic,rombach2022high,ho2020denoising} inherently capture both semantic and detailed visual information through their denoising process.

To incorporate rich visual information into our \ours model, we repurpose diffusion models as decoders within an autoencoder~\cite{hinton1993autoencoders,kingma2013auto,bengio2013generalized,vincent2008extracting} framework. 
\ours embeddings are mapped into the latent space of the T2I diffusion model's vocabulary tokens—specifically, the index-to-embedding lookup matrix of the text encoder~\cite{radford2021learning,raffel2020exploring}, which converts text token IDs into embeddings.
Using diffusion models as decoders for \textit{semantic-level image reconstruction}, rather than traditional VAE decoders~\cite{kingma2013auto} or MAE~\cite{he2022masked} designed for \textit{pixel-level reconstruction}, encourages the model to learn meaningful semantic representations that are highly transferable to diverse visual scene understanding tasks. 
This design enables \ours to harness the rich visual knowledge embedded in diffusion models while maintaining a lightweight structure, making it well-suited for a broad range of understanding tasks beyond diffusion models' original generative applications.

The \ours model consists of a vision encoder that extracts visual representations from the input image and an attention pooling layer that transforms the visual representation into \ourslc tokens.
During training, \ours model is optimized with an image reconstruction loss, receiving gradients from the frozen diffusion model and its text encoder to fine-tune the \ourslc tokens for accurately reconstructing images with similar appearance.
Additionally, we propose the TailDrop strategy during training, where the last $k$ \ourslc tokens are randomly dropped to encourage the earlier tokens to encapsulate richer semantic information. During inference, the number of tokens can be dynamically adjusted to meet user requirements.

Our \ours model is designed to support both image generation and understanding tasks. 
\textbf{\textit{For image generation:}} \ours tokens can be directly used as ``text-prompts'', enabling the re-creation of semantically and visually similar images.
Experiments on COCO image reconstruction demonstrate that our \ours significantly outperforms its counterparts image-guided \dalle3~\cite{betker2023improving} and DeDiffusion~\cite{wei2024diffusion} in terms of the layout, semantic, and
style consistency with the input image, based on human studies.
Notably, even with just a single \ours token, the FID score of \ours remains lower than that of DeDiffusion~\cite{wei2024diffusion}, showcasing the representational power of \ourfull. 
Additionally, \ours embeddings can seamlessly integrate with text prompts, for example, \textit{``an image similar to [ViLex tokens], in Van Gogh style''}, enabling \textbf{\textit{multimodal image generation}} and DreamBooth~\cite{ruiz2023dreambooth} tasks in a zero-shot fashion by prompting a frozen T2I model with both visual and textual inputs.
\textbf{\textit{For image understanding:}} replacing the strong semantic-pretrained backbone SigLIP~\cite{zhai2023sigmoid} with \ours's vision encoder in vision-language models~\cite{beyer2024paligemma} leads to improvements across various vision-language tasks, including image and video captioning~\cite{chen2015microsoft,xu2016msr}, visual question answering~\cite{goyal2017making}, and referring segmentation~\cite{yu2016modeling}.

\par \noindent \textbf{The main contributions of our work are:} 
\begin{itemize}
    \item We propose \ours, an image encoder that maps images into the text space of text-to-image diffusion models. The resulting image embeddings capture both high-level semantics and intricate visual details that are otherwise challenging to convey in natural language.
    \item \ours enables zero-shot unsupervised DreamBooth by prompting T2I models with both \ours tokens and text prompts, without requiring fine-tuning a T2I model or modifying its architectures.
    \ours also improves image reconstruction quality compared to previous baselines for generating semantically similar images, reducing FID by a large margin on MS-COCO using the same token count.
    \item \ours enhances the understanding capability of the image embeddings. 
    Replacing the image encoder in vision-language models with \ours yields improved performance on various visual understanding tasks, including image/video captioning, visual question answering, and image referring segmentation.
\end{itemize}

%% file: sec/2_related.tex
\section{Related Work}
\label{sec:related}

\noindent \textbf{Image Representation Learning} is a fundamental task in computer vision. 
There are two popular approaches:
representing an image with features optimized for visual scene understanding~\cite{deng2009imagenet,wu2018unsupervised,girdhar2023imagebind,tian2020contrastive,wang2021unsupervised,radford2021learning,li2022blip,zhai2023sigmoid,oquab2023dinov2} or with features optimized for high-fidelity image reconstruction~\cite{he2022masked,kingma2013auto,patashnik2021styleclip,yu2024image}, which is often used in image generation~\cite{saharia2022photorealistic,podell2023sdxl}.
Understanding-focused representations like those in CLIP~\cite{radford2021learning}, SigLIP~\cite{zhai2023sigmoid}, DINO~\cite{caron2021emerging}, and DINOv2~\cite{caron2021emerging,oquab2023dinov2} capture high-level semantic information but lose pixel-level details. 
Conversely, reconstruction-focused features, commonly from AutoEncoder-based models (AEs), like VAE~\cite{kingma2013auto}, MAE~\cite{he2022masked}, and BEiT~\cite{bao2022beit}, retain fine image details but often lack semantic richness, limiting their utility in downstream tasks like vision-language modeling~\cite{li2024llava,beyer2024paligemma}.
AutoEncoders, while effective for pixel-level fidelity, often struggle with discriminative tasks. 
Their focus on reconstructing local, semantically agnostic details leads to suboptimal performance in tasks demanding rich, discriminative representations, such as linear evaluation on ImageNet~\cite{bao2022beit,chen2020generative,donahue2019large}. 
We intend to propose a new vision encoder that provides image representations for both image understanding and generation tasks. 

\noindent \textbf{Image Inversion for Diffusion Models.} The goal of image inversion is to determine the text prompt that can be used for generating a specific source image. Prompt-inversion~\cite{mahajan2024prompting,mokady2023null} uses gradient descent to move from the pixel space to the text-embedding space. Techniques like Dreambooth~\cite{ruiz2023dreambooth,ruiz2024hyperdreambooth} and textual-inversion~\cite{gal2022image} learn special text tokens for given instances, but require gradient-based training for each individual image, making them slow at inference time. 
Also, DreamBooth needs to determine the LORA adapters for model architecture changes and is not generic for visual understanding. 
Recently, DeDiffusion~\cite{wei2024diffusion} proposed training a model to generate the inverse text using Gumbel softmax, but the quality of reconstruction is limited by what can be represented by text tokens.
Our approach bypasses discrete language-based text representations, enabling higher-quality reconstructions and efficient single-pass inference.

\noindent \textbf{Representation Learning with Diffusion Models} has been explored by several previous works~\cite{chen2024deconstructing,yang2023diffusion,xu2023open,wang2024diffusion,yu2024spae,hudson2024soda}. For instance, l-DAE~\cite{chen2024deconstructing} uses the diffusion loss as a self-supervised learning objective, while ODISE~\cite{xu2023open} employs diffusion-pretrained features for zero-shot panoptic segmentation. 
DIVA~\cite{wang2024diffusion} shows that finetuning a CLIP backbone~\cite{radford2021learning} with gradients from diffusion models enhances localization capabilities. 
In contrast to these methods, we harness the built-in knowledge of T2I models to learn visual features, effectively framing the generation of text embeddings that reconstruct an image as a powerful representation learning objective.

\noindent \textbf{Image Tokenization}, commonly associated with variational autoencoders (VAEs)~\cite{kingma2013auto}, is crucial for compressing images into a lower-dimensional space for diffusion model training. 
VQVAEs~\cite{rombach2022high,yu2021vector} utilizes a discrete codebook for quantizing latent representations, while recent works like MagViTv2~\cite{yu2023language}, FSQ~\cite{mentzer2023finite}, and BSQ~\cite{zhao2024image} improve quantization with direct binary encoding. 
More recent image tokenizers~\cite{tian2024visual,yu2024image,li2024imagefolder} propose new encoding strategies, including scale prediction~\cite{tian2024visual,li2024imagefolder} and 1D compression~\cite{yu2024image}. 
Unlike these tokenizers, which predict features as noise, our model predicts features conditioned on the diffusion model. Thus, instead of aiming for lossless reconstruction, our focus is on recreating images with high semantic fidelity. 

%% file: sec/3_method.tex
\def\figPipeline#1{
    \captionsetup[sub]{font=small}
    \begin{figure*}[#1]
    \setlength{\belowcaptionskip}{-1pt}
      \centering
      \includegraphics[width=1.0\linewidth]{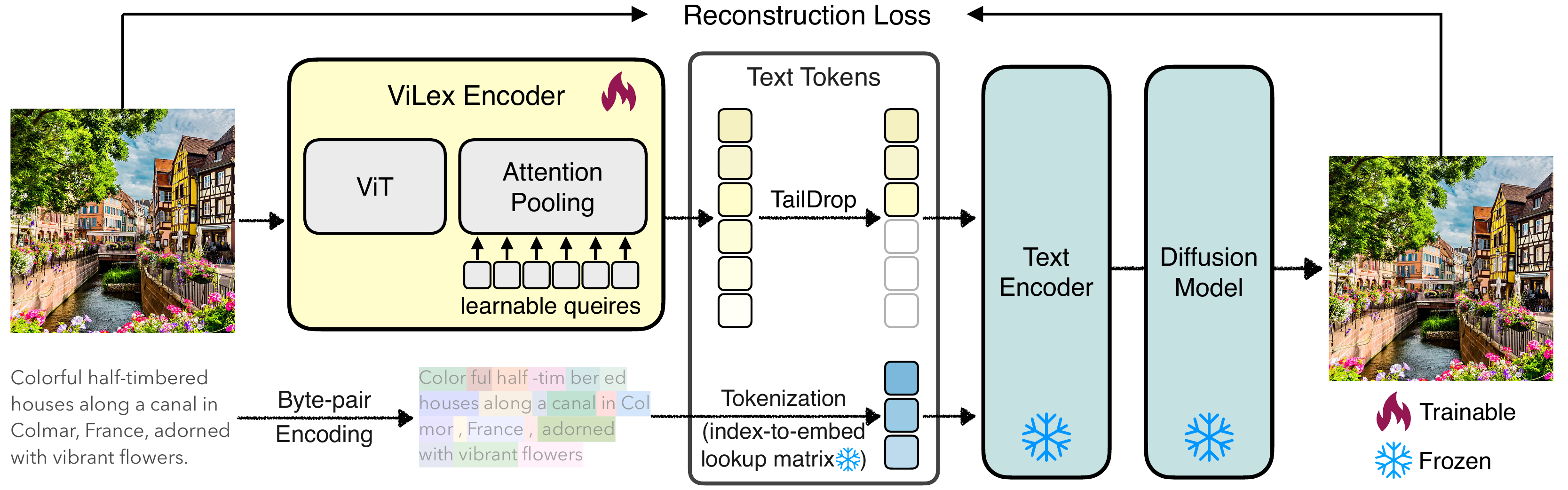}\vspace{-2pt}
      \caption{ 
      \textbf{The pipeline of \ours}: We learn a \ourfull from a frozen diffusion model using an image reconstruction loss. 
      After training, \ours can be directly used as the ``text-prompt'' to a frozen text encoder, \eg, CLIP or T5, enabling the re-creation of semantically similar images \text{without} the need for actual text prompts. 
      In addition, during training, we implement the TailDrop strategy, where the last $k$ tokens are randomly dropped, encouraging earlier tokens in \ours to carry richer semantic information. 
      \ours tokens can be utilized independently as ``text'' tokens for image generation or combined with natural language tokens for prompting T2I diffusion models with both visual and textual cues for multimodal image generation.
      }
      \label{fig:detailed-pipeline}
    \end{figure*}
}

\def\figvsDALLE#1{
    \captionsetup[sub]{font=small}
    \begin{figure*}[#1]
      \vspace{-8pt}
      \centering
      \includegraphics[width=1.0\linewidth]{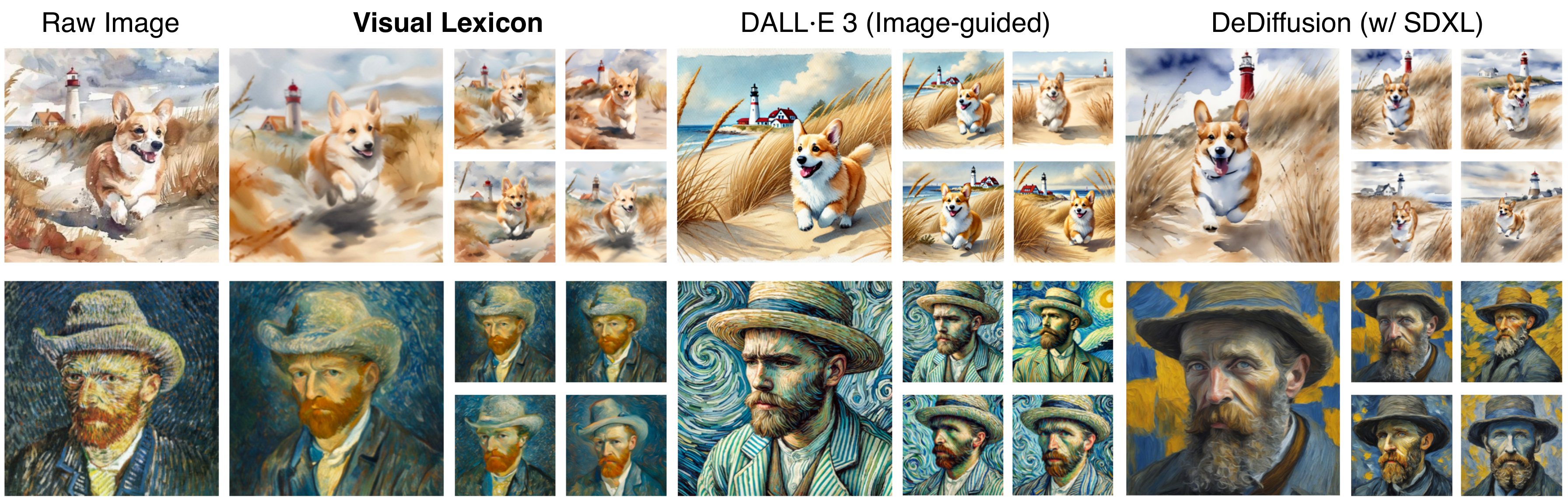}
      \vspace{-18pt}
      \caption{
      \ours retains more visual details in image-to-image generation compared to \dalle3~\cite{betker2023improving} and DeDiffusion~\cite{wei2024diffusion}, accurately capturing elements such as image style (\eg, the oil painting style in row 1), layout (\eg, the relative position of the corgi and the lighthouse), pose (\eg, the corgi’s stance), and object shapes (\eg, the shape of Van Gogh's hat). This enables \ours to produce images that are both semantically and visually consistent with the original input. 
      Even models with text embeddings in a shared language-vision space, like \dalle3, capable of generating semantic variations of an image, struggle to faithfully reconstruct the original appearance of the input image.
      For image-guided \dalle results, we provide the input images along with the text prompt, \textit{``generate an image exactly the same as the input image''}. 
      For DeDiffusion, we follow its official image-to-image generation pipeline and use SDXL~\cite{podell2023sdxl} as the T2I model.}
      \label{fig:image-to-image}
    \end{figure*}
}

\def\figNumTokens#1{
    \captionsetup[sub]{font=small}
    \begin{figure}[#1]
      \centering
      \includegraphics[width=1.0\linewidth]{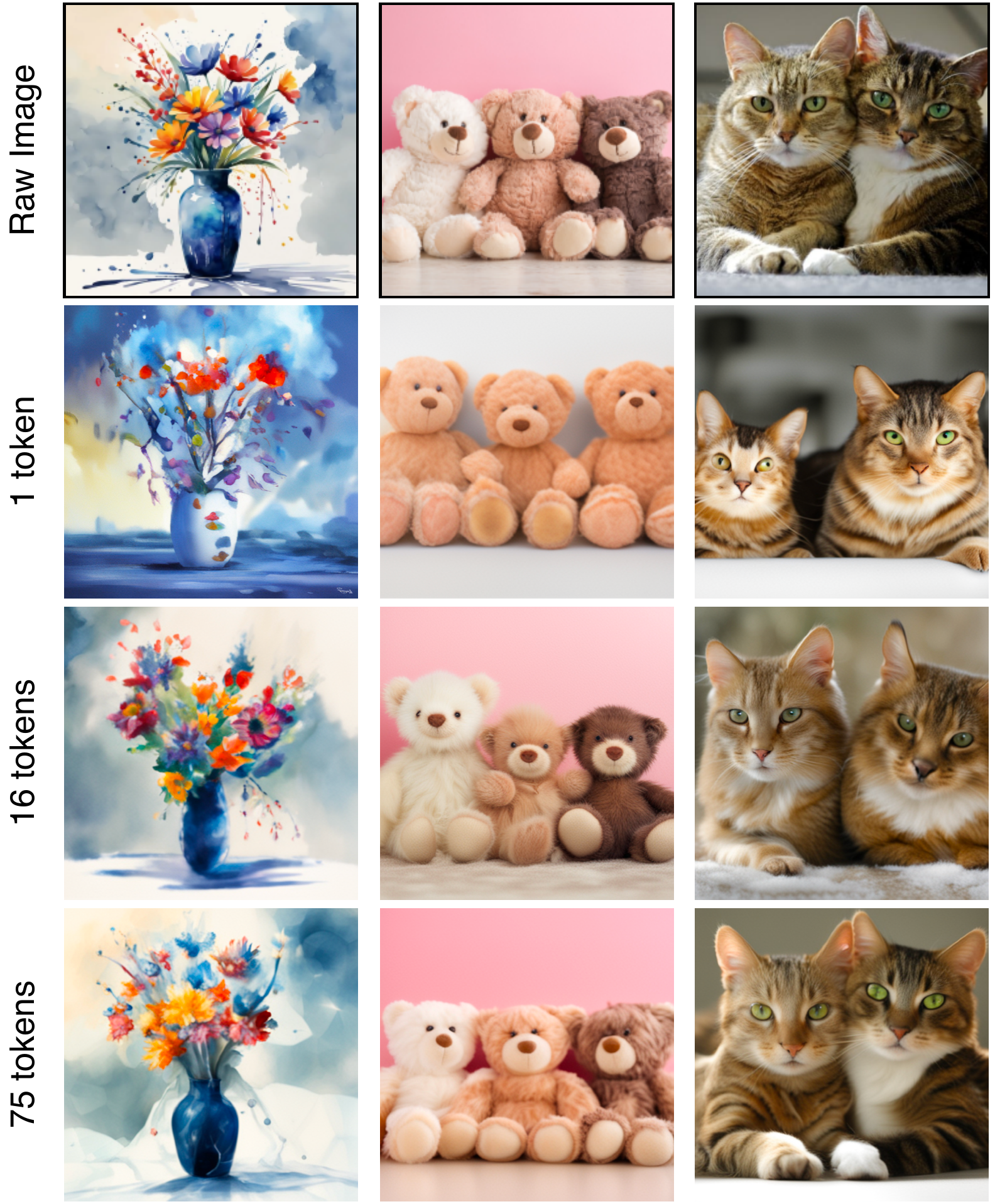}
      \vspace{-18pt}
      \caption{\textbf{Qualitative results of semantic-level image reconstruction with varying numbers of \ours tokens.} 
      \ours tokens enable the reconstruction of semantically similar images even with a single token, effectively capturing high-level semantics such as categories, object counts, and overall poses. 
      As the number of tokens increases to 16, finer details begin to emerge: mid-level features like image styles, colors, and object textures become apparent. 
      With 75 \ours tokens, the reconstructions achieve high appearance fidelity, incorporating fine-grained details that blend both low-level and high-level information, such as precise object shapes, sizes, and cross-instance relationships.}
      \label{fig:num-tokens}
    \end{figure}
}

\def\figvsDreamBooth#1{
    \captionsetup[sub]{font=small}
    \begin{figure*}[#1]
      \setlength{\belowcaptionskip}{-6pt}
      \centering
      \includegraphics[width=1.0\linewidth]{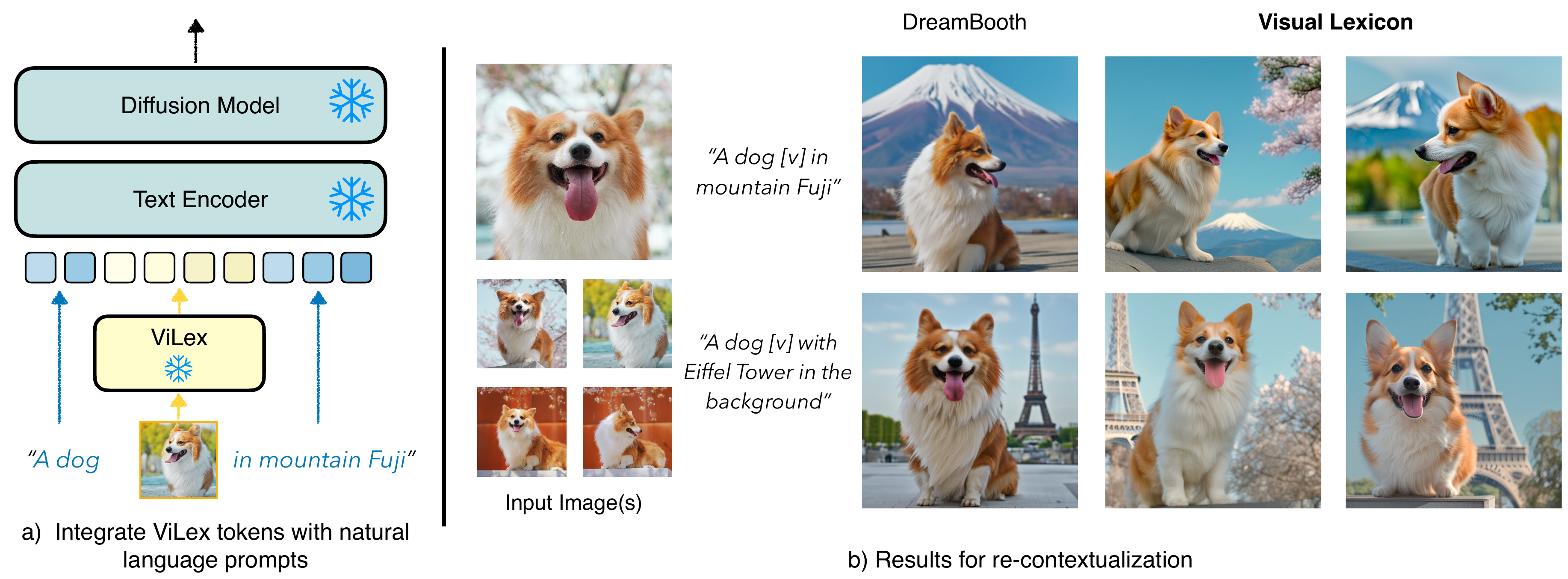}
      \vspace{-16pt}
      \caption{
      \ours can be seamlessly integrated with natural language prompts for \textbf{zero-shot unsupervised image re-contextualization} using a frozen text-to-image (T2I) diffusion model. 
      Unlike DreamBooth~\cite{ruiz2023dreambooth}, \ours requires no fine-tuning of the T2I model on a set of input images from the same object or modifications to the model architecture (\eg, adding a LoRA~\cite{hu2021lora} adapter). 
      Instead, \ours is a universal model that enables zero-shot, unsupervised re-contextualization by simply prompting the T2I model with \ours tokens and corresponding text prompt tokens, just like how we use real words.
      \textbf{a)} The inference pipeline demonstrating image re-contextualization.
      \textbf{b)} Qualitative comparisons with DreamBooth, with DreamBooth results taken from their project page.}
      \label{fig:dreambooth}
    \end{figure*}
}

\def\figvsDreamBoothArt#1{
    \captionsetup[sub]{font=small}
    \begin{figure}[#1]
      \setlength{\belowcaptionskip}{-6pt}
      \centering
      \includegraphics[width=1.0\linewidth]{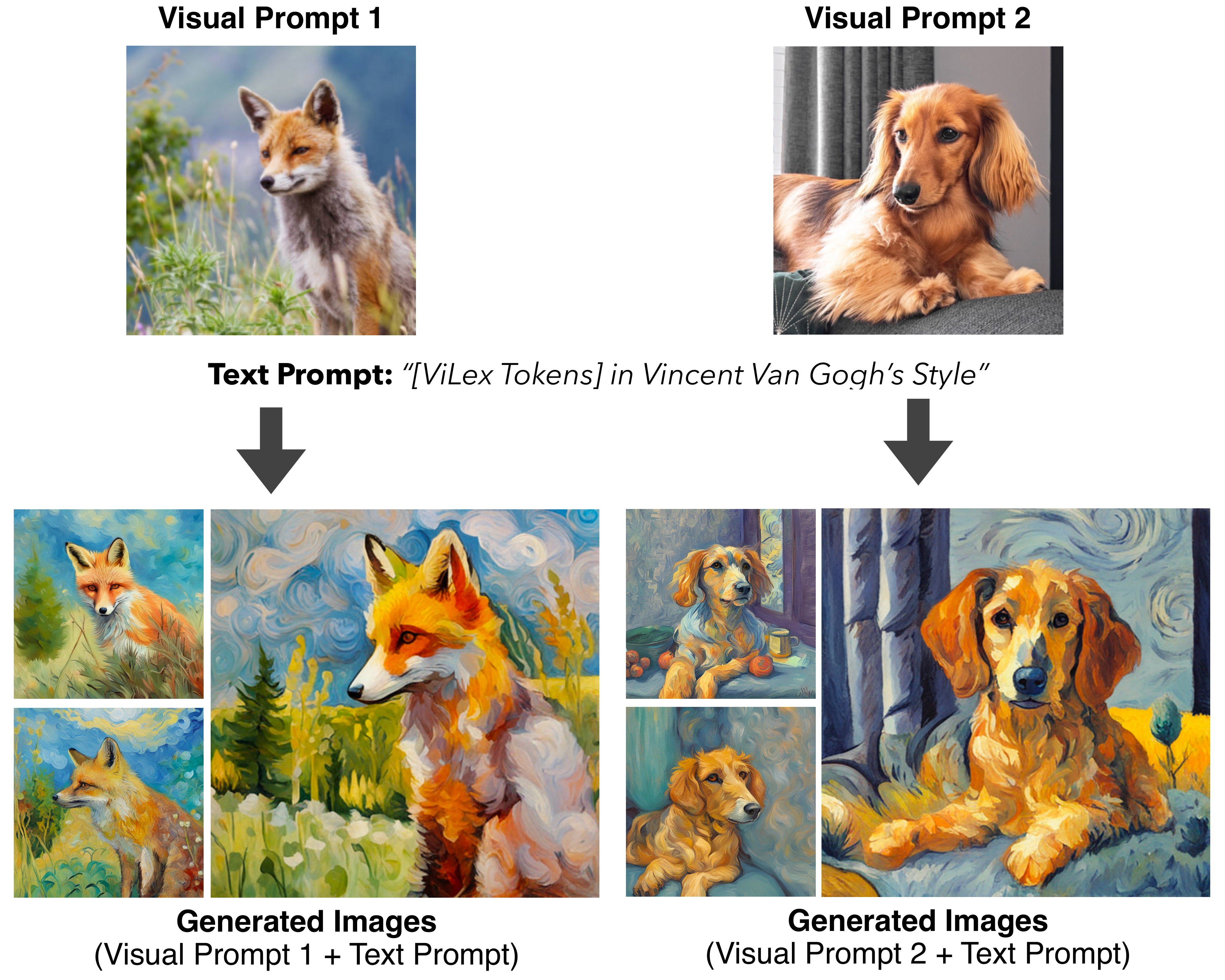}
      \vspace{-16pt}
      \caption{
      \ours can also support zero-shot unsupervised art rendition via prompting T2I models with \ours and text prompts. }
      \label{fig:dreambooth-art}
    \end{figure}
}


\figPipeline{th!}

\section{Describe an Image with a \ourfull}
\label{sec:method}
In this section, we introduce \ours that maps images directly into the text space, while effectively preserving complex visual details that are difficult to express in natural language. 
Our representation effectively acts as a new ``language'' for text-to-image generation and a strong vision encoder for downstream vision-language tasks. 
We will introduce our approach in \cref{sec:methods}, and the technical details in \cref{sec:impl_details}. \cref{fig:detailed-pipeline} presents the overview of \ours.

\subsection{\ourfull} 
\label{sec:methods}
\noindent \textbf{Approach overview}. 
\ours aims to capture high-level semantic representations -- such as object categories and layouts -- while also preserving rich visual details like styles, patterns, and textures that are difficult or even impossible to describe in natural language.

To accomplish this, we train \ours through a self-supervised learning pipeline, with a frozen pretrained text-to-image (T2I) diffusion model serving as the source of supervisory signals. This approach enables \ours to extract visual representations that encompass both semantic-level understanding and intricate visual features.
Unlike previous works that directly use diffusion models as feature extractors~\cite{baranchuk2022labelefficient,li2023your,gan2024instructcv,xu2023open}, we take a different approach.
In our framework, diffusion models act as decoders in an autoencoder~\cite{hinton1993autoencoders,kingma2013auto,bengio2013generalized,vincent2008extracting} pipeline, which allows us to bake the semantic richness learned by these models into a vision encoder.
As a result, \ours benefits from the detailed and rich visual representations of diffusion models while being significantly more lightweight, making it suitable for a broader range of visual scene understanding applications beyond diffusion models' generative origins.

\noindent \textbf{Training approach.} 
As depicted in \cref{fig:detailed-pipeline}, we employ an autoencoder framework to learn \ours from a pre-trained text-to-image diffusion model. 
The overall approach  consists of three key components:
1) Image Encoder: A vision transformer (ViT)~\cite{dosovitskiy2020vit} based image encoder that extracts visual representations from the input image.
2) Image-to-text projection: An attention pooling module that transforms the visual representation into \ours embeddings within the text space. 
These embeddings can be independently used as inputs for the frozen text encoder and the subsequent diffusion model, or concatenated with text tokens derived from natural language.
3) Decoder: A pretrained text-to-image diffusion model serves as the decoder in the autoencoder pipeline, generating images from the ``text'' tokens. 

During training, \ours encoder -- which comprises the ViT and the attention pooling module -- is optimized via the gradients from image reconstruction loss, while the text-to-image (T2I) model remains frozen throughout the process.
After training, \ours can function as a new ``language'', effectively serving as a ``text prompt'' for frozen text encoders such as CLIP~\cite{radford2021learning} or T5~\cite{raffel2020exploring}.
This enables the generation of semantically similar images without traditional text-based prompts, capturing intricate visual details.

\noindent \textbf{Represent images as text embeddings.}
Since \ours is designed to serve as text tokens for text-to-image (T2I) diffusion models, we project the \(k\) patch-level representations \({p}_i\) of an image \(i\) into the text space using a multi-head cross-attention layer~\cite{yucoca}, denoted as \(f(\cdot)\).
This layer contains \(n\) learnable queries and uses the \(k\) output patch tokens from the image encoder as inputs. 
These \(k\) patch tokens function as both keys and values within the cross-attention mechanism.
Through this setup, the model learns to pool the \(k\) patch tokens into \ours embeddings, denoted as \({v}\), consisting of \(n\) tokens such that \({v}_i = f({p}_i)\), as shown in \cref{fig:detailed-pipeline}. 

To illustrate how to implicitly align \ours embeddings \(v\) with text tokens \(c\) compatible with a pretrained T2I model, let's first examine how actual text prompts are tokenized.
Using Byte-Pair Encoding (BPE)~\cite{gage1994new,sennrich2015neural,brown2020language} as an example -- a tokenizer employed by CLIP~\cite{radford2021learning} -- BPE tokenizes text into sub-word units, effectively managing large vocabularies and handling rare or unseen words. The process has two steps:
\textit{1. Tokenization with BPE}: Each input sentence is tokenized using BPE, yielding a sequence of sub-word tokens. For instance, the phrase ``hello world'' might be tokenized as: \(\text{tokens} = [\text{“hel”, “lo”, “wor”, “ld”}]\).
\textit{2. Embedding Lookup}: Each sub-word token is mapped to a learned embedding vector via a pre-trained vocabulary lookup matrix $\mathcal{V}$. If \(e_i\) denotes the embedding for token \(i\), each sub-word token with an index \(t_i\) is given by \(e_i = \mathcal{V}[{t_i}]\). 

The \ours embeddings \(v\) are trained to be implicitly aligned with the latent space of the lookup matrix $\mathcal{V}$, ensuring compatibility with T2I diffusion models.
Before feeding \ours embeddings \(v\) independently or as part of the concatenated token sequence \([v, c]\) to the T2I model, where $c$ is the text tokens, we add \(\text{[BOS]}\) and \(\text{[EOS]}\) tokens at the beginning and end of the sequence, respectively. 

\figvsDALLE{t!}

\noindent \textbf{Text-free guidance for multimodal image generation.}
Classifier-Free Guidance (CFG)~\cite{ho2021classifier} has been a popular technique to enhance the quality of generated samples of diffusion models by controlling the trade-off between adhering to a given prompt and producing diverse outputs. 
In CFG, two sets of samples are generated: one conditioned on the input (\eg, text prompt) and one unconditioned, allowing flexible guidance during the sampling process.
Let $\epsilon_{\theta}(x_t, c)$ denote the noise predicted by the model conditioned on a prompt \(c\) at time \(t\), and $\epsilon_{\theta}(x_t)$ denote the unconditioned noise prediction. In CFG, the final prediction $\epsilon_{\text{guided}}$ is computed as:
\begin{equation}
    \epsilon_{\text{guided}} = \epsilon_{\theta}(x_t) + w \cdot (\epsilon_{\theta}(x_t, c) - \epsilon_{\theta}(x_t))
\end{equation}
Increasing the guidance scale \(w\) intensifies the prompt adherence, improving fidelity to \(c\) at the cost of diversity. 

Inspired by CFG, we introduce Text-Free Guidance (TFG) for our multimodal image generation, balancing the influence of a given text prompt \(c\) and \ours embedding \(v\).
TFG controls the trade-off by incorporating visual representations from \ours alongside the text prompt, enabling finer control over generated images.
In TFG, we modify the noise prediction by combining the conditioned prediction on the visual and text prompts, denoted as \(\epsilon_\theta(x_t, [v, c])\), and the prediction conditioned on \ours alone, \ie \(\epsilon_\theta(x_t, v)\). The TFG noise prediction \(\epsilon_{\text{tfg}}\) is then computed as:
\begin{equation}
\epsilon_{\text{tfg}} = \epsilon_\theta(x_t, v) + w_\text{tfg} \cdot \left( \epsilon_\theta(x_t, [v, c]) - \epsilon_\theta(x_t, v) \right)
\end{equation}
where \(w_\text{tfg}\) is the guidance scale, allowing us to control the impact of \ours tokens relative to the text prompt.
TFG enables multimodal image generation by incorporating both textual and visual cues, \textit{without} requiring T2I model architectural changes or using LoRA~\cite{hu2021lora} adapters.

\noindent \textbf{TailDrop for dynamic visual token compression.}
Different images contain varying amounts of information, this creates a trade-off between representation compactness and detail richness.
We propose a flexible token budget method using a similar masking strategy as in SoundStream~\cite{zeghidour2021soundstream}, which we refer to as TailDrop.
Specifically, during training, we randomly drop the last $k$ \ours tokens.
Since the early tokens are more frequently independently used for image generation, the earlier tokens in \ours are encouraged to carry richer semantic information. 
After the model training, during the inference time, users can dynamically adjust the number of tokens in \ours to suit the needs.

\noindent \textbf{Training loss.} We adopt the standard diffusion~\cite{sohl2015deep,ho2020denoising,song2019generative} training objective to optimize \ours, backpropagating the reconstruction loss to update its parameters. In a diffusion model, the denoising objective aims to learn a model \(\epsilon_\theta(x_t, t)\) that predicts the noise \(\epsilon\) added to data \(x_0\) at timestep \(t\). Given a noisy sample \(x_t\), the objective minimizes the difference between the predicted and true noise:
\begin{equation}
    \mathcal{L}_{\text{denoise}} = \mathbb{E}_{x_0, \epsilon, t} \left[ \|\epsilon - \epsilon_\theta(x_t, t)\|^2 \right],
\end{equation}
where \(x_t = \sqrt{\alpha_t} x_0 + \sqrt{1 - \alpha_t} \epsilon\), with \(\alpha_t\) controlling the noise schedule. This loss enables the model to reverse the diffusion process, gradually reconstructing \(x_0\) from \(x_t\).

\figvsDreamBooth{t!}
\figvsDreamBoothArt{t!}

%% file: sec/4_experiment.tex

\section{Experiments}
\label{sec:experiments}

\label{sec:exp-setup}

\subsection{Technical  Details}
\label{implementation-details}
\label{sec:impl_details}
We describe main technical details here and provide the full details in the supplement.

\noindent \textbf{Text-to-image diffusion model.}
Following DeDiffusion~\cite{wei2024diffusion}, we use Imagen~\cite{saharia2022photorealistic} as the base text-to-image diffusion model, adapting the U-Net architecture from~\cite{ronneberger2015u,nichol2021improved} with 600M parameters, an embedding dimension of 256, and an input resolution of 64×64. 
The text encoder of Imagen is OpenCLIP ViT-H/14~\cite{ilharco_gabriel_2021_5143773,cherti2023reproducible} with a vocabulary size of 49408.
The U-Net conditions on text embeddings via a pooled embedding vector, which is added to the diffusion timestep embedding.
Imagen is further conditioned on the full sequence of text embeddings by incorporating cross-attention over the text embeddings at multiple resolutions.

\noindent \textbf{Model architecture of \ours.} 
\ours consists of two components: a ViT-based image encoder and a transformer-based attention pooling module. 
Both components are unfrozen during the training process.
For the image encoder, we use a pretrained SigLIP@224~\cite{zhai2023sigmoid}.
SigLIP utilizes ViT-base as the backbone and is pretrained on the WebLI dataset~\cite{chen2022pali} using a sigmoid loss and trained on English image-text pairs, with input images resized to 224×224.
The attention pooling module contains \(n\) learnable queries, where \(n\!\leq\!75\), along with [SOS] and [EOS] tokens to ensure the total token count remains within the 77-context length limit defined by the CLIP text encoder~\cite{radford2021learning,ilharco_gabriel_2021_5143773}. The attention pooling layer comprises $5$ transformer blocks, which are always randomly initialized.

\noindent \textbf{Model training.}
The training data is obtained from WebLI~\cite{chen2022pali}, enabling training on either images alone or with image-text pairs.
We found that joint image-text training and our TFG are essential for enabling multimodal image generation. However, training without text captions does not negatively impact performance on downstream vision-language tasks.
Following~\cite{saharia2022photorealistic,wei2024diffusion}, we use Adafactor optimizer~\cite{shazeer2018adafactor} and a weight decay of 0.01. 
Training is performed with a batch size of 2048 over 300K steps, which takes approximately 2.5 days on 64 TPUv5 chips. 
The ViT is initialized with a pretrained SigLIP model and the attention pooling layers are randomly initialized. 
We use learning rate \(1\!\times\!10^{-5}\) for the image encoder and  \(3\!\times\!10^{-4}\) for the attention pooling layers, with a cosine learning rate decay and a 10K-step linear warmup. 
{More training details are in the supplement.}
After training, our \ours encoder maps an image to \ours representations. We next evaluate two capabilities of these frozen \ours representations: image generation and visual understanding. 

\subsection{Experiments on Image Generation}

\hide{
To assess \ours's ability to preserve semantic information and visual details in images, we evaluate its performance using Fréchet Inception Distance (FID)~\cite{heusel2017gans} and Inception Score (IS)~\cite{szegedy2016rethinking} scores. We use De-Diffusion~\cite{wei2024diffusion} and \dalle3~\cite{betker2023improving} as strong baselines, following the same pipeline of image-to-text-to-image.
We also compare the performance of \ours against De-Diffusion across various token counts without using any ground-truth text prompts. 
Besides the FID and IS scores, we conducted a human evaluation on the image-to-image generation results, assessing model performance in terms of semantic consistency (\eg, all specified instances in the input images are present), layout consistency (\eg, arrangement of foreground objects and background scenes), and style consistency (\eg, alignment of image styles, colors, textures, etc.).
For qualitative analysis of multimodal image generation, we input both natural language text prompts and \ours tokens into the T2I model Imagen, applying our text-free guidance (TFG) during inference with a guidance scale \(w_{\text{tfg}} = 10\).
}

\label{sec:exp-results}
\hide{
\noindent \textbf{Image-to-image generation.}
We evaluate \ours's image-to-image generation capabilities by providing \ours tokens as inputs to a frozen text-to-image diffusion model.
The goal is to recreate visually similar images based on the \ours representation of an input image.
Compared to other models like DeDiffusion~\cite{wei2024diffusion} or \dalle3~\cite{betker2023improving}, \ours demonstrates superior fidelity in capturing fine details such as style, object positioning, and visual characteristics, as confirmed through qualitative analysis in \cref{fig:image-to-image} and FID scores in \cref{tab:fid}. 
Our human evaluations in \cref{tab:humam-eval} also highlight improvements in semantic consistency, layout alignment, and style preservation, making \ours a compelling solution for high-fidelity image-to-image generation. 
}

\tabFID{t!}
\tabHumanEval{t!}
\tabSigLIP{t!}

\noindent \textbf{Image-to-image generation} aims to generate similar images given an input image.
Feeding our extracted \ours features on the input image to our T2I model~\cite{saharia2022photorealistic} with different diffusion random noise yields re-created images.
We compare with two popular models: \dalle3~\cite{betker2023improving} and DeDiffusion~\cite{wei2024diffusion}, both of which convert the input image to explicit texts.
DeDiffusion~\cite{wei2024diffusion} used exactly the same encoder architecture and T2I model as us, giving us an apple-to-apple comparison between our \ours feature and texts.
We conducted both human studies (\cref{tab:humam-eval}) and quantitative metrics on Fréchet Inception Distance (FID)~\cite{heusel2017gans} and Inception Score (IS)~\cite{szegedy2016rethinking} scores (\cref{tab:fid}).

The quantitative results in \cref{tab:fid} show that our \ours efficiently preserves visual information, surpassing the explicit text counterpart~\cite{wei2024diffusion} even with only 1 token in both FID and IS metrics. 
We further evaluate the effectiveness of \ours through human assessments on image-to-image generation tasks.
Results in \cref{tab:humam-eval} indicate that \ours significantly outperforms the baselines, achieving 98\% win rates in layout alignment, 95\% in semantic fidelity, and 98\% in style preservation against~\cite{wei2024diffusion}. \ours also outperforms \dalle3.
We present qualitative comparisons in \cref{fig:image-to-image}.

\figNumTokens{t!}

\noindent \textbf{Zero-shot unsupervised multimodal image generation {with identity preserving}.}
One advantage of mapping images to the word space is to embed images directly into a sentence in an interleaved way.
This enables multimodal image generation using a pure text-to-image model without finetuning.
In \cref{fig:dreambooth}, we demonstrate the ability of \ours to serve as a ``language'' for multimodal image generation while preserving the object identity in a given image.
The conventional approach to this task like DreamBooth~\cite{ruiz2023dreambooth} requires LoRA~\cite{hu2021lora} adapters and a test-time finetuning with a set of images to learn an embedding for object identity which is slow and computationally expensive, while ours directly infers the target identity feature using our encoder.
As shown in \cref{fig:dreambooth,fig:dreambooth-art}, \ours improves the resulting images by embedding detailed semantic and visual context, producing coherent multimodal results that reflect both the textual prompt and the visual cues provided.
We highlight again that our model is not finetuned for this task and does not require a text-time-finetuning like DreamBooth, but simply prompting a standard T2I model with \ours embeddings and corresponding text prompts. 

\noindent \textbf{When do we need more \ours tokens?} 
\cref{fig:num-tokens} presents qualitative results of semantic-level image reconstruction with varying numbers of \ours tokens, illustrating the gradual refinement of visual details as the token count increases. With just 1 token, \ours effectively captures high-level semantic information, such as object categories, counts, and poses. As the the token count increases to 16, mid-level features, including image styles, colors, and textures, start to emerge, enhancing the overall visual representation. Finally, with 75 tokens, \ours captures fine-grained details such as precise object shapes, sizes, and intricate cross-instance relationships. For instance, in the third-row image, the two cats' poses are accurately reconstructed, including their specific positioning and interaction as they cuddle together. These results demonstrate \ours' adaptability, providing users with the flexibility to balance semantic richness and visual detail based on the application requirements.

\subsection{Experiments on Image Understanding}

We next verify that our frozen \ours features can be directly used for understanding tasks, by feeding them to a large language model~\cite{team2024gemma}.
We use PaliGemma~\cite{beyer2024paligemma} as our visual-language model (VLM) architecture.
PaliGemma~\cite{beyer2024paligemma} is an open-source VLM with a SigLIP-So400m~\cite{zhai2023sigmoid} vision encoder and a Gemma-2B~\cite{team2024gemma} language model.
To use \ours in PaliGemma, we replace the SigLIP~\cite{zhai2023sigmoid} vision encoder with our \ours encoder.
In all our following experiments, we don't finetune our \ours encoder, and only finetune the following large language model~\cite{team2024gemma} to adapt to the tasks following the PaliGemma transfer design~\cite{beyer2024paligemma}.

\noindent \textbf{Improving vision encoders over SigLip.} 
We first conduct a comparison between the visual encoder learned from our \ours pretraining and the popular SigLip~\cite{zhai2023sigmoid}.
To ensure a fair comparison in terms of the architecture and the number of tokens, we use our features right after the ViT-encoder without the pooling layer.
\cref{tab:siglip} shows the results.
We also show the image reconstruction FID.
The results show that \ours feature consistently improves the strong SigLip model by over $1$ point margin on a variety number of vision-language tasks, including image captioning, referring segmentation, and video understanding.
This shows that \ours can effectively upgrade existing vision encoders to perform better in both reconstruction and understanding.

\tabMainCompare{t!}

\noindent \textbf{Vision-language tasks.} 
\ours can also serve as a strong vision encoder for vision-language models. 
\ours tokens are inherently ready to use for vision-language frameworks, therefore, they can be directly utilized without requiring fine-tuning the image encoder. 
Despite freezing the image encoder—including both the ViT and attention pooling layers—we observed significant performance improvements across various tasks, such as visual question answering, image captioning, and referring expression segmentation, as demonstrated in \cref{tab:res-concat} and \cref{tab:vqa}.
Following prior works like QwenVL-7B, LLaVA1.5-13B, LLaVA1.6-13B~\cite{bai2023qwen,liu2024improved,liu2024llava}, we adopt a multi-grid input strategy for extracting visual representations to ensure a fair comparison with these methods.
However, unlike these methods, which increase the overall number of tokens by 2$\sim$5 times, we use only 16 \ours tokens per grid. This results in only 25\% increase in token count (adding just 80 tokens) while achieving substantial performance improvements and setting new state-of-the-art results on multiple VQA benchmarks.

\tabVQA{t!}

\noindent \textbf{Pixel-reconstruction or semantic-reconstruction?} 
Although VAE~\cite{kingma2013auto,hinton1993autoencoders,rombach2022high} excels at preserving pixel-level details during image reconstruction, its~\cite{rombach2022high} lack of semantic richness leads to substantially poorer performance (often $>$8 times lower) in vision-language modeling compared to \ours, as demonstrated in \cref{tab:res-concat}. 

%% file: sec/5_summary.tex
\section{Conclusions}
We introduce \ours, a \ourslc that maps images directly into the text space, while effectively preserving complex visual details that are difficult to express in natural language. Our representation can be seamlessly integrated into text prompts from natural language for both multimodal image generation and downstream vision-language tasks. \ours can also improve both image understanding and reconstruction capabilities of pretrained vision encoders by fine-tuning them using \ours's training approach.

\noindent \textbf{Acknowledgment.} We sincerely thank Alexei A. Efros, Ren Ng, David Minnen, Lijun Yu, Xiuye Gu, Haiwen (Haven) Feng, Renhao Wang, Baifeng Shi, and Tony Long Lian for their insightful discussions and valuable feedback on our paper.
We also thank Chen Wei for her assistance in reproducing the baseline results.

%% file: sec/suppl.tex
\def\tabSigLIPAppendix#1{
    \begin{table*}[#1]
    \tablestyle{2.2pt}{1.02}
    \small
    \begin{center}
    \begin{tabular}{lcHcccclcccccclcccccclcc}
    &&& \multicolumn{5}{c}{\text{Image Captioning}} && \multicolumn{6}{c}{{Visual Question Answering}} && \multicolumn{6}{c}{Image Segmentation} && Video \\
    \cline{4-8} \cline{10-15} \cline{17-22} \cline{24-24}
    Backbone & \#steps & FID ($\downarrow$) & {\rotatebox{80}{COCOcap}} & {\rotatebox{80}{COCO-35L}} & {\rotatebox{80}{TextCap}} & {\rotatebox{80}{SciCap-Val}} & {\rotatebox{80}{SciCap-Test}} &&
    {\rotatebox{80}{VQAv2-Val}} & \rotatebox{80}{TextVQA} & {\rotatebox{80}{OKVQA}} & {\rotatebox{80}{SciQA}} & {\rotatebox{80}{VizWizQA}} & {\rotatebox{80}{GQA}} &&
    \rotatebox{80}{RC-val} & \rotatebox{80}{RC-testA} & \rotatebox{80}{RC-testB} & \rotatebox{80}{RCp-testA} & \rotatebox{80}{RCp-testB} & \rotatebox{80}{RCg-test} &&
    {\rotatebox{80}{MSRVTT}} \\ [.1em]
    \Xhline{0.8pt}
    \hline
    Original SigLIP & - &    2.54 &    139.7 &    138.6 &    122.1 &    131.7 &    135.5 && 81.4 &    51.9 &    57.1 &    85.9 &    74.3 &    64.8 &&    66.2 &    69.0 &    63.6 &    63.3 &    55.3 &    59.6 &&    69.4 \\
    \rowcolor{gray!15}
    \ours SigLIP    & 150k &\bf - &\bf 140.5 &\bf 138.8 &\bf 122.3 &\bf 132.6 &\bf 135.5 &&\bf 81.4    &\bf 52.1 &\bf 57.3 &\bf 86.1 &\bf 74.5 &\bf 65.1 &&\bf 66.5 &\bf 69.3 &\bf 64.2 &\bf 64.1 &\bf 55.6 &\bf 60.2 && \bf 70.6 \\
    \rowcolor{gray!15}
    \rowcolor{gray!15}
    \ours SigLIP    & 600k &\bf 2.38 &\bf 141.5 &\bf 140.0 &\bf 124.0 &\bf 134.3 &\bf 136.1 &&\bf 81.8    &\bf 52.7 &\bf 58.3 &\bf 89.3 &\bf 75.0 &\bf 65.4 &&\bf 67.5 &\bf 69.7 &\bf 65.6 &\bf 65.2 &\bf 57.2 &\bf 62.6 && \bf 71.4 
    \end{tabular}
    \end{center}\vspace{-6pt}
    \caption{
    {\ours improves both image understanding and reconstruction capabilities of vision encoders} by fine-tuning them using \ours's training approach. 
    \textbf{Extending the fine-tuning of SigLIP with the \ours approach from 150k to 600k steps results in improved overall model performance across evaluated benchmarks.}
    We use PaliGemma's~\cite{beyer2024paligemma} framework for linear evaluation, replacing the vision encoder with either the fine-tuned SigLIP in \ours or the official one, and freeze vision encoder and fine-tune the model on downstream tasks.
    We use the same hyper-parameters and model architecture for a fair comparison.}
    \label{tab:siglip-appendix}
    \end{table*}
}

\def\figHumanStudy#1{
    \captionsetup[sub]{font=small}
    \begin{figure*}[#1]
      \centering
      \includegraphics[width=1.0\linewidth]{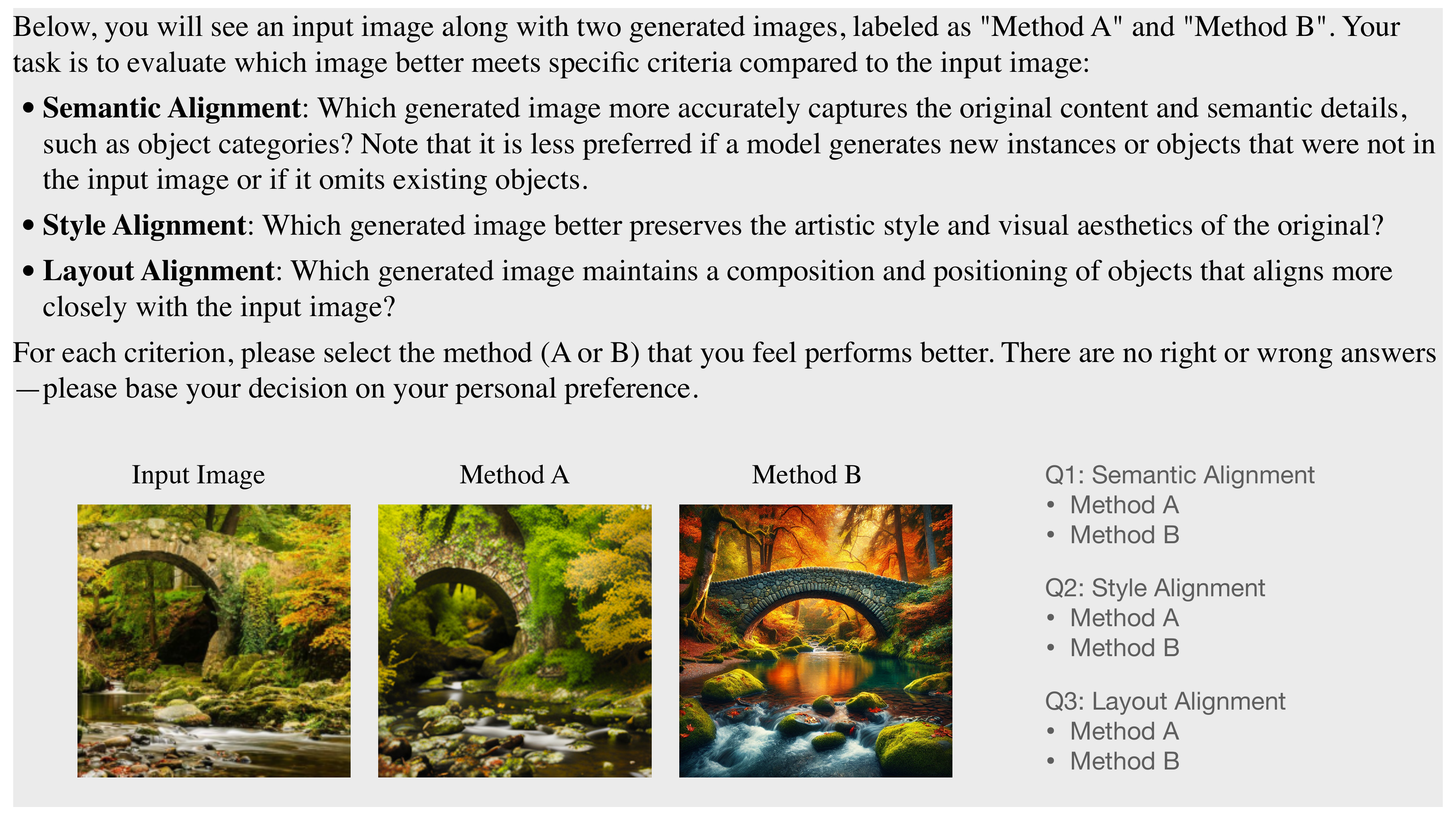}\vspace{-2pt}
      \caption{The instructions and question format used for human study.
      }
      \label{fig:human-study}
    \end{figure*}
}

\def\figDemoAppendix#1{
    \captionsetup[sub]{font=small}
    \begin{figure*}[#1]
      \centering
      \includegraphics[width=1.0\linewidth]{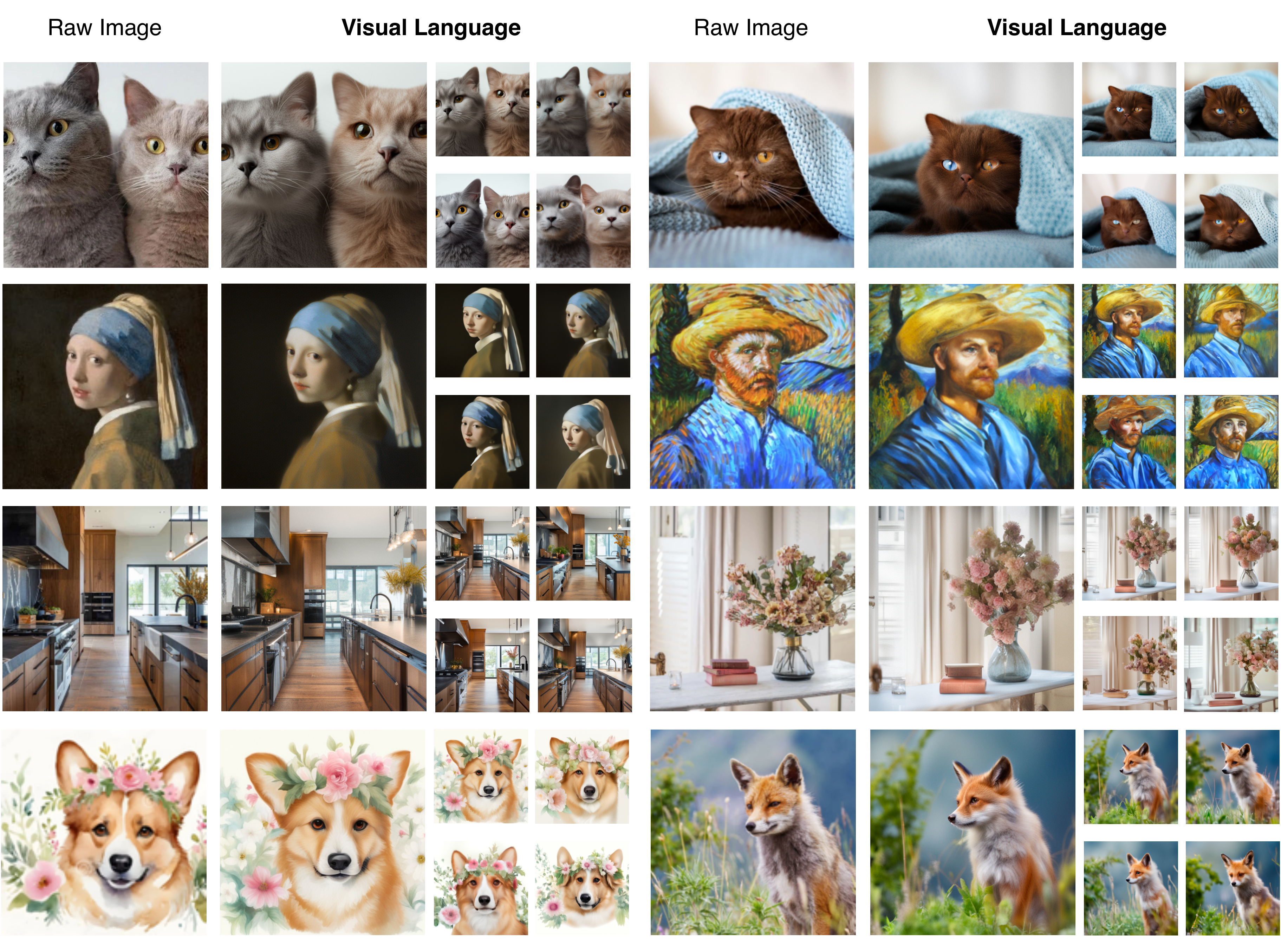}\vspace{-2pt}
      \caption{ 
      More demo results of generating a set of images (generated under different diffusion noises), which are highly semantically and visually similar to each other, by using \ours tokens as ``text'' prompts for text-to-image diffusion models. 
      }
      \label{fig:demo-appendix}
    \end{figure*}
}

\def\figStyleTransfer#1{
    \captionsetup[sub]{font=small}
    \begin{figure*}[#1]
      \centering
      \includegraphics[width=1.0\linewidth]{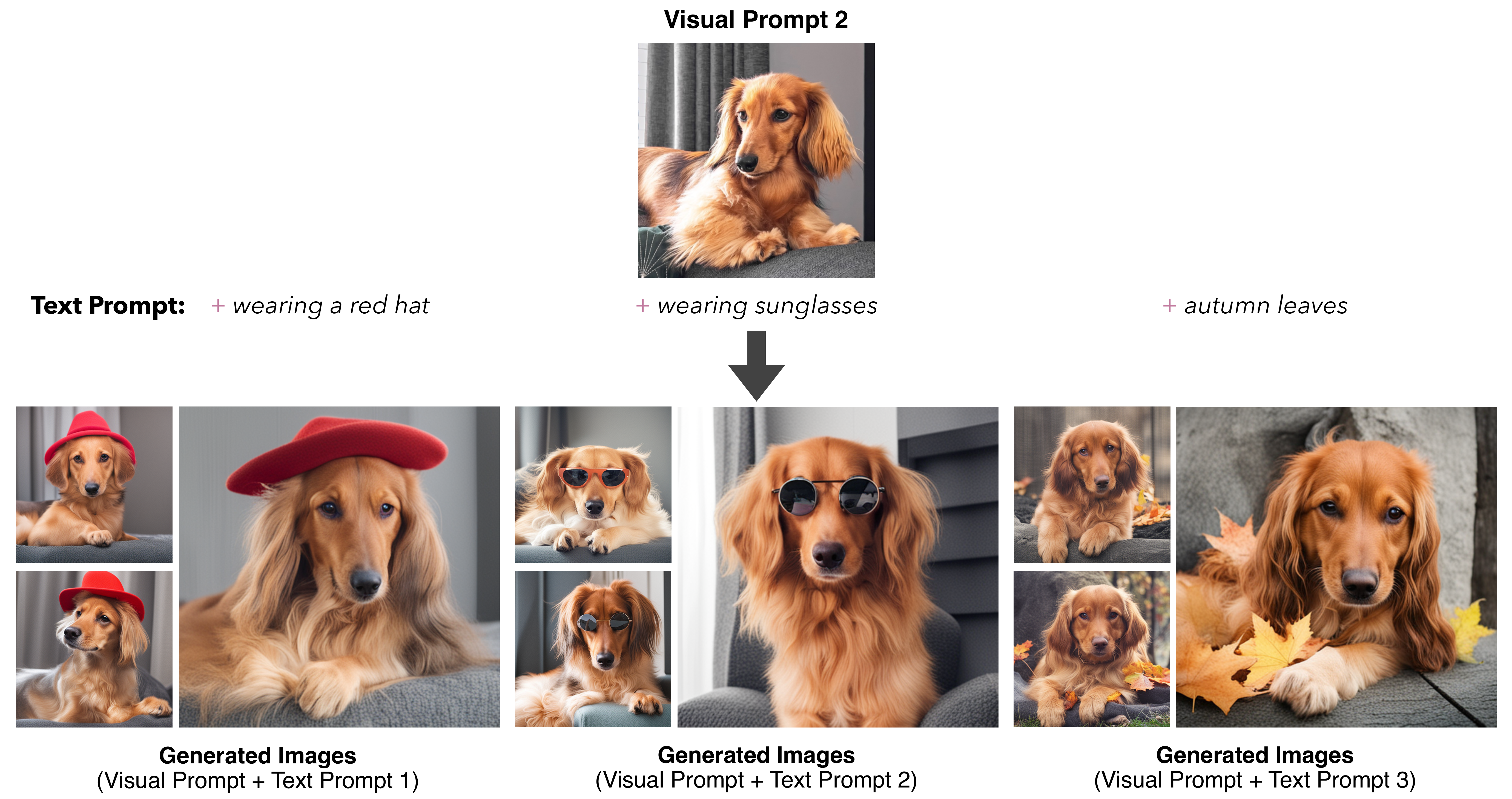}\vspace{-2pt}
      \caption{ 
      More demo results of zero-shot accessorization via prompting a frozen text-to-image generation model with our visual prompts (\ie, \ours tokens) and text prompts from natural language. 
      }
      \label{fig:demo-style}
    \end{figure*}
}

\def\tabNumBlocks#1{
    \begin{table}[#1]
    \tablestyle{5pt}{1.02}
    \small
    \begin{center}
    \begin{tabular}{cccc}
    \#layers & FID & COCOCaps \\ [.1em]
    \Xhline{0.8pt}
    \hline 
    2 & 2.62 & 140.7 \\
    \rowcolor{gray!15}
    5 & 2.58 & 141.5 \\
    8 & 2.52 & 141.0 \\
    \end{tabular}
    \end{center}\vspace{-12pt}
    \caption{Ablation study on number of attention pooling layers.}
    \label{tab:num-attn-blocks}
    \end{table}
}

\def\tabVisionEncoder#1{
    \begin{table}[#1]
    \tablestyle{3.8pt}{1.02}
    \small
    \begin{center}
    \begin{tabular}{ccc|cc}
    Datasets & CoCa & F.T. w/ \ours & SigLIP & F.T. w/ \ours \\ [.1em]
    \Xhline{0.8pt}
    \hline 
    COCOCaps   & 131.6 & \bf 135.8 & 139.7 & \bf 141.5 \\
    MSRVTTCaps & 56.1  & \bf 60.2  & 69.4  & \bf 71.4  \\
    \end{tabular}
    \end{center}\vspace{-12pt}
    \caption{
    Fine-tuning vision encoders with the \ours approach enhances image understanding performance across various pretrained models, including CoCa~\cite{yucoca} and SigLIP~\cite{zhai2023sigmoid}.
    F.T. denotes fine-tuning the vision encoder, such as CoCa, during the \ours model pretraining stage. 
    }
    \label{tab:various-vision-encoders}
    \end{table}
}

\clearpage
\renewcommand{\thefigure}{A\arabic{figure}}
\setcounter{figure}{0}
\renewcommand{\thetable}{A\arabic{table}}
\setcounter{table}{0}
\renewcommand{\thesection}{A\arabic{section}}
\setcounter{section}{0}

\maketitlesupplementary

\section{Technical Details}

We introduced the main technical and implementation details of our \ours model in the main paper, here we provide a more comprehensive explanation. 

\noindent \textbf{Text-to-image diffusion model.}
Following DeDiffusion~\cite{wei2024diffusion}, we use Imagen~\cite{saharia2022photorealistic} as the base text-to-image diffusion model, adapting the U-Net architecture from~\cite{ronneberger2015u,nichol2021improved} with 600M parameters, an embedding dimension of 256, and an input resolution of 64×64. 
The text encoder of Imagen is OpenCLIP ViT-H/14~\cite{ilharco_gabriel_2021_5143773,cherti2023reproducible} with a vocabulary size of 49408.
The U-Net conditions on text embeddings via a pooled embedding vector, which is added to the diffusion timestep embedding.
Imagen is further conditioned on the full sequence of text embeddings by incorporating cross-attention over the text embeddings at multiple resolutions. 
The Imagen model uses \(v\)-prediction~\cite{salimansprogressive} as its objective, with a batch size of 2048, and is trained for 3 million steps. 
As a baseline model, Imagen achieves an FID of $6.52$ on 30K 64×64 MS-COCO 2014 validation images~\cite{saharia2022photorealistic}.
During image generation inference, we use a super-resolution model, such as an SDXL upsampler, to upsample the image resolution from 64×64 to 512×512 for better visualizations. 

\noindent \textbf{Model architecture of \ours.} 
\ours consists of two components: a ViT-based image encoder and a transformer-based attention pooling module. 
Both components are unfrozen during the training process.
For the image encoder, we use a pretrained SigLIP-So400M@224~\cite{zhai2023sigmoid}.
SigLIP utilizes ViT-base as the backbone and is pretrained on the WebLI dataset~\cite{chen2022pali} using a sigmoid loss and trained on English image-text pairs, with input images resized to 224×224.
The model architecture of the ViT-base is shape-optimized on 400M training samples for improving the model efficiency and speed. 
In our method, the attention pooler is implemented as a single multi-head attention layer with learnable queries, using the encoder output as both keys and values. This allows the attention pooling module to effectively aggregate embeddings of varying lengths.
The attention pooling module contains \(n\) learnable queries, where \(n\!\leq\!75\), along with [SOS] and [EOS] tokens to ensure the total token count remains within the 77-context length limit defined by the CLIP text encoder~\cite{radford2021learning,ilharco_gabriel_2021_5143773}. The attention pooling layer comprises $5$ transformer blocks, which are always randomly initialized. 

\noindent \textbf{Model training.}
The training data is obtained from WebLI~\cite{chen2022pali}, enabling training on either images alone or with image-text pairs.
We found that joint image-text training and our TFG are essential for enabling multimodal image generation. However, training without text captions does not negatively impact performance on downstream vision-language tasks. 
Following~\cite{saharia2022photorealistic,wei2024diffusion}, we use Adafactor optimizer~\cite{shazeer2018adafactor} and a weight decay of 0.01. 
Training is performed with a batch size of 2048 over 300K steps, which takes approximately 2.5 days on 64 TPUv5 chips. 
We found that double the training steps (from 300k to 600k) can further improve the model performance on increasing the performance of a pretrained vision encoder. 
The ViT is initialized with a pretrained SigLIP model and the attention pooling layers are randomly initialized. 
We use learning rate \(1\!\times\!10^{-5}\) for the image encoder and  \(3\!\times\!10^{-4}\) for the attention pooling layers, with a cosine learning rate decay and a 10K-step linear warmup, and a weight decay of 0.01.
After training, our \ours encoder maps an image to \ours representations. We next evaluate two capabilities of these frozen \ours representations: image generation and visual understanding. 

\tabSigLIPAppendix{t}
\figHumanStudy{t!}

\figDemoAppendix{t!}
\figStyleTransfer{t!}

\noindent \textbf{PaliGemma experiments.}
To evaluate the effectiveness of the proposed \ours approach in enhancing a pretrained vision encoder for vision-language tasks, we integrate our vision encoder into the PaliGemma~\cite{beyer2024paligemma} framework and replace the vision encoder with either the fine-tuned SigLIP-So400M~\cite{zhai2023sigmoid} from ViLex or the official version without model fine-tuning, freezing the vision encoder and fine-tuning the model on downstream tasks. Following PaliGemma's official pipeline, we transfer the model to a variety of individual academic benchmarks using a unified transfer approach with minimal hyperparameter tuning.
To ensure fair comparison, we applied the same hyperparameter sweeping strategy for both the baseline and our fine-tuned vision encoder, reporting the best results for each. 
This structured approach allows us to fairly assess the impact of the proposed \ours method on a wide range of vision-language tasks.
The sweeping parameters for these tasks are as follows:
COCOCap~\cite{lin2014microsoft} (COCO image captioning task) and COCO-35L~\cite{thapliyal2022crossmodal} (COCO captions translated in 35 languages): learning rate (4e-6, 5e-6, 6e-6), epochs (5, 10), dropout (0, 0.02, 0.05).
TextCaps~\cite{sidorov2020textcaps} (image captioning with reading comprehension): learning rate (4e-6, 6e-6), and training epochs (5, 10).
For SciCaps~\cite{hsu2021scicap} (captions for scientific figures): learning rate (6e-5, 7e-5), dropout (0.1, 0.2), and label smoothing (0.1, 0.2).
For VQAv2~\cite{goyal2017making} (visual question answering): label smoothing (0.0, 0.1), dropout (0.0, 0.1), and weight decay (0, 1e-6). 
For TextVQA~\cite{singh2019towards} (visual reasoning based on text in images): learning rate (4e-6, 6e-6).
For OKVQA~\cite{marino2019ok} (outside knowledge VQA), ScienceQA~\cite{lu2022learn} (science question answering), and VizWizVQA~\cite{gurari2018vizwiz} (VQA from people who are blind): learning rate (8e-6, 1e-5), and dropout (0.0, 0.02).
For GQA~\cite{hudson2019gqa} (VQA on image scene graphs): learning rate (5e-6, 1e-5), and dropout (0.0, 0.02, 0.05).
For RefCOCO~\cite{kazemzadeh2014referitgame,yu2016modeling,mao2016generation} (referring expression segmentation): label smoothing (0.1, 0.2), epochs (60, 100), and dropout (0, 0.05).
For MSRVTT-Caps~\cite{xu2016msr} (open-domain short video captioning): weight decay (0, 1e-6), dropout (0, 0.2), and epochs (20, 40). 

\section{Human Study}
We conduct human studies to evaluate the quality of generated images using an image-to-image pipeline, focusing on three criteria: Semantic Alignment, Style Alignment, and Layout Alignment.
For Semantic Alignment, participants judge which generated image more accurately captures the original content and semantic details, such as object categories. Introducing new instances or omitting existing ones from the input image is considered less desirable.
For Style Alignment, participants assess which generated image best retains the artistic style and visual aesthetics of the original.
For Layout Alignment, participants evaluate which generated image maintains a composition and positioning of objects that closely matches the input image.

The results of this evaluation are reported in \cref{tab:humam-eval} of the main paper. Detailed instructions and the question format for the human study are shown in \cref{fig:human-study}.

\section{Ablation Study}
\noindent \textbf{Training Steps.} We observed that extending the fine-tuning steps of the vision encoder using our \ours pipeline leads to improved performance across nearly all evaluated benchmarks, as shown in \cref{tab:siglip-appendix}. Specifically, increasing the training steps from 150k to 300k yields significant gains. Further extending the training to 600k steps provides marginal improvements compared to the 300k-step results. 
The largest improvements are observed in datasets that demand stronger spatial understanding, such as the referring expression segmentation datasets RefCOCO/+/g. 

\noindent \textbf{Number of attention pooling layers.} 
Although increasing the number of attention pooling layers improves image reconstruction performance (as indicated by a lower FID score), it also introduces a trade-off with image understanding capabilities. 
As shown in \cref{tab:num-attn-blocks}, we found that using 5 attention pooling layers provides the optimal balance between image generation quality and developing an effective vision encoder for visual scene understanding.
\tabNumBlocks{h!}

\noindent \textbf{Vision encoders.} 
The \ours approach effectively enhances various vision encoders for downstream visual scene understanding tasks. We initialize the vision encoder of \ours with either the CoCa~\cite{yucoca} pretrained ViT or the SigLIP~\cite{zhai2023sigmoid} pretrained ViT-So400M.
Similar to our experiments in previous sections, 
We observed consistent performance improvements for both image understanding tasks, such as COCOCaps~\cite{lin2014microsoft}, and video understanding tasks, such as MSRVTT-Caps~\cite{xu2016msr}. Compared to a roughly 2\% improvement for SigLIP in terms of CIDEr score on COCOCaps, the gains for CoCa were even more substantial, reaching approximately 4\%.
The flexibility to consistently improve different pretrained models demonstrates \ours's generalizability across various types of vision encoders. 
\tabVisionEncoder{h!}

\section{Demo Results}

\noindent \textbf{Semantic-level image reconstruction.} 
In this section, we present additional demo results in \cref{fig:demo-appendix}, showcasing a set of images generated with varying diffusion noises and different random seeds. These images demonstrate high semantic and visual consistency, leveraging \ours tokens as ``text'' prompts for text-to-image diffusion models. 
However, as shown in the results, our model occasionally misses small objects in the scene. This limitation primarily stems from using a low-resolution text-to-image diffusion model as the base during the \ours model's pretraining phase. We hypothesize that this issue could potentially be mitigated by employing a higher-resolution T2I model as the base model.

\noindent \textbf{Prompting a frozen T2I model with both visual and textual prompts.} 
In the main paper, we have demonstrated that \ours tokens can serve as a novel visual ``language'' for multimodal image generation. 
Unlike methods such as DreamBooth~\cite{ruiz2023dreambooth,ruiz2024hyperdreambooth} and textual inversion~\cite{gal2022image}, which require: (1) learning specialized text tokens for specific instances, (2) gradient-based training for each individual image, and (3) the use of LORA adapters~\cite{hu2021lora} to modify the model architecture, DreamBooth must be fine-tuned separately for each object (or each set of images corresponding to the same object).
In contrast, \ours enables several DreamBooth tasks like image re-contextualization, artistic rendition and accessorization, as illustrated in \cref{fig:demo-style}, \cref{fig:dreambooth} and \cref{fig:dreambooth-art}, by simply prompting a frozen T2I model with a combination of our visual prompts (\ie, \ours tokens) and natural language text prompts. This approach does not require changes to the architecture of a pretrained text-to-image generation model or any fine-tuning of the T2I model itself. All tasks are performed in a zero-shot and unsupervised manner.